\batchmode
\documentclass{article}
\usepackage{latexsym, amssymb, amsfonts}
\usepackage[dvips]{graphicx}
\usepackage{fancyvrb}

\title{Syntax, Parsing and Production of Natural Language
in a Framework of Information Compression by
Multiple Alignment, Unification and Search\protect\footnote{{\em Journal of Universal Computer Science} 6(8), 781--829, 2000.}}

\author{J Gerard Wolff\\
\small (University of Wales, Bangor, UK\\
gerry@sees.bangor.ac.uk)}

\begin{document}

\maketitle

\begin{abstract}

This article introduces the idea that {\it information compression by
multiple alignment, unification and search} (ICMAUS) provides a
framework within which natural language syntax may be represented in a
simple format and the parsing and production of natural language may be
performed in a transparent manner.

In this context, {\it multiple alignment} has a meaning which is
similar to its meaning in bio-informatics but with significant
differences, while {\it unification} means a simple merging of matching
patterns, a meaning which is related to but simpler than the meaning of
that term in logic. The concept of {\it search} in the present context
means search for alignments which are `good' in terms of information
compression, using heuristic methods or arbitrary {\it constraints} (or both)
to restrict the size of the search space.

These concepts are embodied in a software model, SP61. The organisation
and operation of the model are described and a simple example is
presented showing how the model can achieve parsing of natural
language.

Notwithstanding the apparent paradox of `decompression by compression',
the ICMAUS framework, without any modification, can produce a sentence
by decoding a compressed code for the sentence. This is illustrated
with output from the SP61 model.

The article includes four other examples - one of the parsing of a
sentence in French and three from the domain of English auxiliary
verbs. These examples show how the ICMAUS framework and the SP61 model
can accommodate `context sensitive' features of syntax in a relatively
simple and direct manner.

An important motivation for this research is the possibility of
developing the ICMAUS framework as a unifying framework for diverse
aspects of computing in addition to those described in this article. Other
aspects which appear to fall within the scope of the ICMAUS framework but
which are outside the scope of this article, include the representation
of natural language semantics, best-match pattern recognition and
information retrieval, deductive and probabilistic reasoning, planning
and problem solving, and unsupervised inductive learning.

\end{abstract}


{\bf Category:} I.2.7

\section{Introduction}

This article introduces the idea that {\it information compression} (IC) 
{\it by multiple alignment, unification and search} (ICMAUS) provides a
framework within which natural language syntax may be represented in a
simple format and the parsing and production of natural language may be
performed in a transparent manner.

In this context, {\it multiple alignment} has a meaning which is
similar to its meaning in bio-informatics but with significant
differences, while {\it unification} means a simple merging of matching
patterns, a meaning which is related to but simpler than the meaning of
that term in logic. In the present context, {\it search} means search
for alignments amongst patterns which are `good' in terms of
information compression, with {\it constraints} to reduce the size of
the search space as described in Section
\ref{multiple_alignment_problems}.

The mechanisms for heuristic search which are incorporated in the
ICMAUS framework as it has been developed here allow syntactic
knowledge to be expressed as {\it patterns} (as described in Section
\ref{representing_grammar}), a mode of expression which is
significantly different from existing formalism.\footnote{In this research, the term {\it pattern} has been adopted as a general term
which means an array of symbols of one, two or more dimensions.
Notwithstanding the fact that this research has so far been largely
restricted to one-dimensional sequences of symbols, the term {\it pattern} is
generally used in preference to the term {\it sequence} as a reminder
of the intention, later in the research programme, to generalise the
concepts to two or more dimensions.  Formal definitions of terms like
{\it symbol} and {\it pattern} as they are used in this research are
given in Appendix \ref{appendix_A}.} The use of patterns (in the sense
of this article), has potential advantages compared with existing
formalisms.

These concepts are embodied in a software model, SP61. This article
describes the organisation and operation of the model with examples of
what the model can do.

\subsection{Novelty of proposals}

Aspects of these proposals which appear to be novel are described in
the following subsections.

\subsubsection{Parsing (with choices at many levels) as multiple alignment}

The most novel feature of the present proposals appears to be the idea
that parsing, in the sense understood in theoretical and computational
linguistics and natural language processing, may be understood as
multiple alignment.

A concept of parsing is already well-established
in the literature on data compression (see, for example, \cite{r28}).
In that context, it means a process of analysing data into segments,
each of which is replaced by a relatively short `code' associated with
the given segment in a `dictionary' of segments.

But this kind of parsing is simpler than `linguistic' kinds of parsing.
In the first case, although segments may have internal hierarchical
structure, alternatives can be chosen only at one level. In
the second kind of parsing, which is the focus of interest in this
article, there may be alternatives at {\it arbitrarily many levels} in
the grammar which is used to guide the parsing.

\subsubsection{Parsing as information compression}

\sloppy Research on parsing and related topics within computational
linguistics and AI does not normally consider these topics in terms of
information compression (IC) (but see, for example, \cite{r7, r18}).
However, there is a well-developed tradition of parsing and linguistic
analysis in terms of probabilities, with associated concepts such as
`stochastic grammars', `maximum-likelihood', `Bayesian inference' and
`statistical analysis' (see, for example, \cite{r1, r8, r13, r16, r29,
r47, r21}) and there is a close connection between probabilities and
IC.\footnote{Measures of frequency or probability have a key role in
techniques for economical coding such as the Huffman method or the
Shannon-Fano-Elias (S-F-E) method (see \cite{r11}). Conversely,
measures of compression may be translated into measures of probability
(see \cite{r34}).}

\subsubsection{Production of language as ICMAUS}

Essentially the same points as were made above about parsing apply also
to the production of language. It is interesting that the ICMAUS framework,
without any modification, lends itself to the production of language as
well as it does to parsing (see Section \ref{decoding_by_compression}).

\subsubsection{Representing syntax with patterns}

As we shall see, the ICMAUS framework allows natural language syntax to
be represented with patterns in a manner which is
significantly different from other formalisms.

\subsection{Background and context}\label{background_and_context}

The proposals in this article have been developed within a programme of
research developing the `SP' conjecture\footnote{Information
compression may be interpreted as a process of maximising {\it
Simplicity} in information (by reducing redundancy) whilst retaining as
much as possible of its non-redundant descriptive {\it Power}. Hence
the sobriquet `SP' which has been applied to these ideas.} that {\it
all kinds of computing and formal reasoning may usefully be understood
as information compression by pattern matching, unification and
search}, and developing a `new generation' computing system based on
this thinking (\cite{r44} to \cite{r32a}).

This entire programme of research is based on an earlier programme of
research into unsupervised learning of language structures (see
\cite{r45, r46} and earlier articles cited there). That research and
the present research are based on principles of Minimum Length Encoding
(MLE). Relevant sources are cited in Section \ref{related_research}.

The overall aim of this research programme is the {\it integration} and
{\it simplification} of concepts in computing and cognition. Besides the
aspects of natural language processing considered in this article, the
ICMAUS framework appears to have potential to accommodate several other
aspects of computing and cognition, including unsupervised learning
\cite{r36}, the representation of non-linguistic `semantic' structures
(examples may be found in \cite{r34, r34c}), mathematics and logic
\cite{r32a}, probabilistic reasoning \cite{r34, r36}, best-match
information retrieval (\cite{r39}) and best-match pattern recognition
(\cite{r37}). It can be argued (\cite{r33}) that the ICMAUS framework
provides an interpretation for the organisation and operation of any
Universal Turing Machine (UTM), and equivalent models of `computing'
such as the Post Canonical System (PCS).

\subsection{Related research}\label{related_research}

As an attempt to integrate concepts across several areas of computing,
the SP programme naturally has many connections with other research in
the several areas that it seeks to integrate. Some connections are
described in \cite{r43, r42, r38}.

In terms of {\raggedright theoretical foundations, the closest links are
with work on Algorithmic Information Theory (AIT, see, for
example, \cite{r19}) and Minimum Length Encoding (MLE, see,
for example, \cite{r26, r32, r25, r6, r15}) which is itself closely
related to Bayesian inference (see, for example, \cite{r10, r22}).

}

\subsubsection{Distinctive features of the SP programme compared with
MLE and AIT}

Although the SP theory is based on MLE principles, there are important
differences in objectives and orientation between the SP programme and
other research in MLE and AIT. These differences are described in
Section 3.6 of \cite{r38} and Sections 7.1 and 7.2 of \cite{r40}).  In
brief, the main differences are:

\begin{itemize}
 
\item The SP programme seeks to integrate {\it all} kinds of computing
and formal reasoning within a framework of information compression. This
goal is broader than it is in other research in AIT or MLE.

\item The SP programme is based on the hypothesis that {\it all} kinds
of information compression may be understood in terms of multiple
alignment, unification and search. In essence this means the hypothesis
that all kinds of information compression is achieved by the
unification of matching patterns. All existing and projected SP models
are restricted to ICMAUS mechanisms and avoid `arithmetic coding' and
other mathematical techniques which are used for information
compression.

The restriction has been imposed in the interests of simplicity in the
SP theory. The aim is to build a theoretical framework from a `bedrock'
of apparently primitive operations of matching symbols and patterns and
unifying symbols and patterns. The theory should avoid including any
concepts that cannot be derived from this foundation.

If, as conjectured, arithmetic and, perhaps, mathematics, may be
understood in terms of ICMAUS (see \cite{r32a}), then compression
techniques that use mathematical concepts may also be understood in
terms of ICMAUS.  But until this has been demonstrated, the adoption of
arithmetic coding or any other mathematical technique would add
unwanted complexity to the SP model.

\item In the first point above, the phrase ``all kinds of computing''
includes the concept of `computing' itself in its full depth and
generality. Thus the SP programme hypothesises that other models of
computing such the Turing model or the Post Canonical System may be
understood in terms of the ICMAUS concepts (see \cite{r33}). By
contrast, researchers in AIT and MLE accept the Turing model (and
equivalent models) as the foundation of concepts in computing.

\end{itemize}

\subsection{Scope of this article}

This paper describes new concepts in the representation of syntax and
in the parsing and production of language. It does not describe a
complete working system, including a complete grammar of one or more
languages, and it should not be evaluated as such.

In the space available, it is possible only to present these proposals
in outline. It has been necessary to omit many details and there are
many associated issues which could not be discussed.

Although integration of concepts right across the field of computing
provides the main motivation for this programme of research (as noted
in Section \ref{background_and_context}), topics other than syntax,
parsing and production of natural language will not be considered
except briefly where they are relevant.

\subsection{Presentation}

In what follows, I have tried to bring the important ideas into relief
by describing them relatively briefly in the body of the article and
moving details into appendices. The main sections after this one are
these:

\begin{enumerate}

\item[\ref{multiple_alignment_problems}] introduces
multiple alignment problems in general terms and describes how the concept
has been generalised in this programme of research.

\item [\ref{parsing_as_multiple_alignment}] describes the use of {\it
patterns} (defined in Appendix \ref{appendix_A}) to represent the broad
features of the syntax of natural language and describes how the
parsing of language may be seen as multiple alignment. 

\item [\ref{MA_compress}] describes in outline how an alignment may
be evaluated in terms of IC.

\item [\ref{the_SP61_model}] describes the main features of the SP61
model, a partial realisation of the ICMAUS framework, running on a
conventional computer.

\item [\ref{decoding_by_compression}] describes how the production of
language may be seen in terms of multiple alignment - between a `coded'
representation of a sentence and rules in a grammar.

\item [\ref{context_sensitive_aspects}] presents a selection of other
examples showing how the ICMAUS framework can accommodate `context
sensitive' features of syntax.

\item [\ref{discussion_conclusion}] discusses briefly some associated
issues and makes some concluding remarks.

\end{enumerate}

The appendices are as follows:

\begin{enumerate}

\item[\ref{appendix_A}] provides formal definitions of the main terms
used in this article.

\item[\ref{appendix_B}] supplements Section
\ref{MA_compress} with a more detailed account of the
method for evaluating the IC associated with any alignment which is
used in the SP61 model.

\item[\ref{appendix_C}] describes the organisation of the SP61 model in
more detail than Section \ref{the_SP61_model}.

\end{enumerate}

Generally speaking, small examples have been used in this article for
the sake of clarity and to save space. It should not be assumed that
the examples represent the limits of what the system can do (see
Section \ref{computational_complexity}).

\section{Multiple alignment
problems}\label{multiple_alignment_problems}

{\it Multiple alignment} is a term borrowed from bio-informatics where
it means the arrangement of two or more sequences of symbols in
horizontal rows one above the other so that, by judicious `stretching'
of sequences where necessary, symbols that match each other from one
sequence to another can be brought into alignment in vertical columns.

A `good' alignment is, in general, one where there is a relatively
large number of {\it hits} (positive matches) between symbols and where any
{\it gaps} (sequences of unmatched symbols) between hits are relatively few
and relatively short. The meaning of `good' in this context is described
in Section \ref{MA_compress} and Appendix \ref{appendix_B}.

Multiple alignments like these are normally used in the computational
analysis of (symbolic representations of) sequences of DNA bases or
sequences of amino acid residues as part of the process of elucidating
the structure, functions or evolution of the corresponding molecules.
An example of an alignment of DNA sequences is shown in Figure \ref{DNA}.

\begin{figure}[b!hpt]
\centering
\begin{BVerbatim}
  G G A     G     C A G G G A G G A     T G     G   G G A
  | | |     |     | | | | | | | | |     | |     |   | | |
  G G | G   G C C C A G G G A G G A     | G G C G   G G A
  | | |     | | | | | | | | | | | |     | |     |   | | |
A | G A C T G C C C A G G G | G G | G C T G     G A | G A
  | | |           | | | | | | | | |   |   |     |   | | |
  G G A A         | A G G G A G G A   | A G     G   G G A
  | |   |         | | | | | | | |     |   |     |   | | |
  G G C A         C A G G G A G G     C   G     G   G G A
\end{BVerbatim}
\caption{\small A `good' alignment amongst five DNA sequences.}
\label{DNA}
\normalsize
\end{figure}

\subsection{Search and the need for
constraints}\label{search_and_constraints}

In this area of research, it is widely recognised that, with the
exception of alignments of patterns which are very small and very few,
the number of possible alignments of symbols is too large to be
searched exhaustively. For any set of patterns of realistic size, a
search which has acceptable speed and acceptable scaling properties can
only be achieved if some kind of {\it constraint} is used:

\begin{itemize}

\item Arbitrary parts of the search space may be excluded {\it a
priori}.  For example, in multiple alignment problems, an upper limit
may be set to the size of any `gap' between `hits', as described
above.

\item With `heuristic' techniques, searching is done in stages, with a
progressive narrowing of the search space in successive stages using
some kind of measure of `goodness' to guide the search. Heuristic
techniques include `hill climbing' (sometimes called `descent'), `beam
search', `genetic algorithms', `simulated annealing', `dynamic
programming' and others. These techniques may be described generically
as `metrics-guided search'.

\end{itemize}

Either or both of these kinds of constraint may be applied. Given one
or both of these kinds of constraint, it is not possible to guarantee
that, for any set of patterns, the best possible alignment has been
found. For many tasks, this guarantee is not necessary and it is
sufficient to find alignments that are ``good enough''.

There is now a fairly large literature about methods for finding good
alignments amongst two or more sequences of symbols. All of them use
constraints of one kind or another and, for that reason, none of them
can guarantee that the best possible result is always found. Some of
the existing methods are reviewed in \cite{r30, r5, r9, r12}. For
reasons which will be explained in the next section, none of the
current methods seem to be entirely suitable for incorporation in the
proposed SP system.

\subsection{Development of the concept of multiple alignment in the
present research}\label{varieties_of_MA}

In this research, concepts associated with multiple alignment and the
multiple alignment concept itself have been adapted and developed in
the following way:

\begin{itemize}

\item One (or more) of the patterns of symbols to be aligned has a
special status and is designated as `New'. In the context of parsing,
this would be the sentence (or other sequence of symbols) which is to
be parsed.

\item All other patterns are designated as `Old'. In the context of
parsing, this would be the patterns of symbols which represent
grammatical `rules' (more about this in Section
\ref{parsing_as_multiple_alignment}, below).

\item A `good' alignment is one which, through the unification of
symbols in New with symbols in Old, and through unifications amongst
the symbols in Old, leads to a relatively large amount of compression
of New in terms of the sequences in Old. How this may be done is
explained in outline in Section \ref{MA_compress} and in
more detail in Appendix \ref{appendix_B}.

\item By contrast with `multiple alignment' as normally understood in
bio-informatics, any given sequence in Old may appear two or more times
in any one alignment and, in these cases, it is possible for the given
sequence to be aligned with itself.

\item As noted already, it is envisaged that, at some point in the
future, the concept of multiple alignment as it is understood here will
be generalised to alignments of patterns with two dimensions or higher
(diagrams, pictures and so on).

\end{itemize}

Notice that two or more appearances of a pattern in an alignment are
repeated {\it appearances} of a single entity in the alignment - and this is
{\it not} the same as having two or more {\it copies} of a given
pattern in an alignment. In the latter case, it is permissible to form
a hit between a given symbol in one copy of a pattern and the
corresponding symbol in another copy. In the case of two or more {\it
appearances} of a pattern in an alignment, it is {\it not} permissible
to form a hit between a symbol in one appearance and the corresponding
symbol in another appearance - because this would mean forming a hit
between one symbol and itself.

\section{Syntax as `patterns' and parsing as multiple
alignment}\label{parsing_as_multiple_alignment}

This section describes how the simpler aspects of syntax may be
represented with patterns and how the parsing of a sentence in terms of
a grammar may be seen in terms of multiple alignment.

The example considered in this section and again in Section
\ref{decoding_by_compression} may give the impression that the ICMAUS
framework is merely a trivial variation of familiar concepts of
context-free phrase-structure grammar (CF-PSG) with their well-known
inadequacies for representing and analysing the `context sensitive'
structures found in natural languages. The examples presented in
Section \ref{context_sensitive_aspects} show that the ICMAUS framework
is much more `powerful' than CF-PSGs and can accommodate quite subtle
context-sensitive features of natural language syntax in a simple and
elegant manner.

\subsection{Representing a grammar with patterns of
symbols}\label{representing_grammar}

Figure \ref{grammar_1} shows a simple CF-PSG describing a fragment of
the syntax of English. This grammar generates sentences like `t h i s b
o y l o v e s t h a t g i r l', `t h a t b o y h a t e s t h i s g i r
l', and so on.  Any of these sentences may be parsed in terms of the
grammar giving a labelled bracketing like this:

\begin{center}
\begin{BVerbatim}
(S(NP(D t h i s)(N b o y))(V l o v e s)
     (NP(D t h a t)(N g i r l)))
\end{BVerbatim}
\end{center}

\noindent or an equivalent representation in the form of a tree.

Figure \ref{grammar_2} shows the grammar from Figure \ref{grammar_1}
expressed as a set of strings, sequences or {\it patterns} of symbols
(as defined in Appendix \ref{appendix_A}). Each pattern in this
`grammar' is like a re-write rule in the CF-PSG notation except that
the rewrite arrow has been removed, some other symbols have been
introduced (`0', `1' and symbols with an initial `\#' character) and
there is a number to the right of each rule.

\begin{figure}[b!hpt]
\centering
\begin{BVerbatim}
S -> NP V NP
NP -> D N
D -> t h i s
D -> t h a t
N -> g i r l
N -> b o y
V -> l o v e s
V -> h a t e s
\end{BVerbatim}
\caption{\small A CF-PSG describing a fragment of English syntax.}
\label{grammar_1}
\normalsize
\end{figure}

The number to the right of each rule in Figure \ref{grammar_2} is a
frequency of occurrence of the rule in a (`good') parsing of a notional
sample of the language. These frequencies have a role in determining
the IC associated with any alignment but their main significance
(considered in \cite{r34} and outsid the scope of this article) is in
determining probabilities associated with any given alignment.

The reasons for the symbols which have been added to each rule will
become clear but a few words of explanation are in order here. The
symbols `0' and `1' have been introduced to differentiate the two
versions of the `D' patterns, and likewise for the `N' patterns and `V'
patterns. They enter into matching and unification in exactly the same
way as other symbols. Although the symbols are the same as are used in
other contexts to represent numbers they do not have the meaning of
numbers in this grammar.

\begin{figure}[b!hpt]
\centering
\begin{BVerbatim}
S NP #NP V #V NP #NP #S (500)
NP D #D N #N #NP (1000)
D 0 t h i s #D (600)
D 1 t h a t #D (400)
N 0 g i r l #N (300)
N 1 b o y #N (700)
V 0 l o v e s #V (650)
V 1 h a t e s #V (350)
\end{BVerbatim}
\caption{\small The grammar from Figure \ref{grammar_1} recast as
patterns of symbols.}
\label{grammar_2}
\normalsize
\end{figure}

The symbols which begin with `\#' (e.g., `\#S', `\#NP') serve as
`termination markers' for patterns in the grammar. Although their
informal description as `termination markers' suggests that these
symbols are meta symbols with special meaning, they have no hidden
meaning and they enter into matching and unification like every other
symbol.

In general, all the symbols that can be seen in Figure \ref{grammar_2}
and other examples in this article are simply `marks' that can be
discriminated from each other by yes/no matches but otherwise have no
intrinsic meaning. Although some of these symbols can be seen to serve
a distinctive role, there is no hidden meaning attached to any of them
and no formal distinction between upper- and lower-case letters or
between digit symbols and alphabetic symbols and so on (see Appendix
\ref{appendix_A}).

\subsection{Parsing as alignment of a sentence and rules in a
grammar}\label{parsing_as_alignment}

Figure \ref{basic1} shows how a parsing of the sentence `t h i s b o y
l o v e s t h a t g i r l' may be seen as an alignment of patterns
which includes the sentence and relevant rules from the grammar shown
in Figure \ref{grammar_2}. The similarity between this alignment and
the conventional parsing may be seen if the symbols in the alignment
are `projected' on to a single sequence, thus:

\begin{center}
\begin{BVerbatim}
S NP D 0 t h i s #D N 1 b o y #D #NP V 0 l o v e s #V
     NP D 1 t h a t #D N 0 g i r l #N #NP #S
\end{BVerbatim}
\end{center}

In this projection, the two instances of `NP' in the second column of
the alignment have been merged or `unified' and likewise for the two
instances of `D' in the third column and so on wherever there are two
or more instances of a symbol in any column.

\begin{figure}[b!hpt]
\centering
\begin{BVerbatim}
0          t h i s        b o y            l o v e s    0
           | | | |        | | |            | | | | |     
1          | | | |        | | |            | | | | |    1
           | | | |        | | |            | | | | |     
2          | | | |        | | |            | | | | |    2
           | | | |        | | |            | | | | |     
3          | | | |        | | |            | | | | |    3
           | | | |        | | |            | | | | |     
4          | | | |        | | |        V 0 l o v e s #V 4
           | | | |        | | |        |             |   
5 S NP     | | | |        | | |    #NP V             #V 5
    |      | | | |        | | |     |                    
6   |  D 0 t h i s #D     | | |     |                   6
    |  |           |      | | |     |                    
7   NP D           #D N   | | | #N #NP                  7
                      |   | | | |                        
8                     N 1 b o y #N                      8
                                                                
0        t h a t        g i r l           0
         | | | |        | | | |            
1        | | | |    N 0 g i r l #N        1
         | | | |    |           |          
2 NP D   | | | | #D N           #N #NP    2
  |  |   | | | | |                  |      
3 |  D 1 t h a t #D                 |     3
  |                                 |      
4 |                                 |     4
  |                                 |      
5 NP                               #NP #S 5
                                           
6                                         6
                                         
7                                         7
                                           
8                                         8
 \end{BVerbatim}
\caption{\small The best alignment found by SP61 with `t h i s b o y l o v e s
t h a t g i r l' in New and the grammar from Figure \ref{grammar_2} in Old.}
\label{basic1}
\normalsize
\end{figure}

This projection is the same as the conventional parsing except that `0'
and `1' symbols are included, right bracket symbols (`)') are replaced
by `termination markers' and each of the upper-case symbols is regarded
both as a `label' for a structure and as a left bracket for that
structure.

Notice that the pattern `NP D \#D N \#N \#NP' appears twice in the alignment
in Figure \ref{basic1}, in accordance with what was said in Section
\ref{varieties_of_MA}. In general, any pattern in the grammar used for
parsing may appear two or more times in an alignment. Other examples
will be seen later.

As was noted in Section \ref{varieties_of_MA}, the sentence or other
sequence of symbols to be parsed is regarded as New, while the rules in
the grammar are regarded as Old. For the sake of readability and ease
of interpretation, New is normally placed at the top of each alignment
with patterns from Old below it.

For the sake of clarity in Figure \ref{basic1} and other alignments
shown in this article, each appearance of a pattern in any alignment is
given a line to itself (so that the two appearances of `NP D \#D N \#N
\#NP' in Figure \ref{basic1} are on two different lines). Apart from
the convention that New is always at the top, the order in which
patterns appear (from top to bottom of the alignment) is entirely
arbitrary. An alignment in which the patterns appear in one order is
totally equivalent to an alignment in which they appear in any other
order, provided all other aspects of the alignment are the same.

All the examples of parsing by alignment shown in this article are
output from the SP61 model and in every case, the alignment shown is
the best alignment (in terms of IC) that the model has found with the
given sentence in New (in row 0) and the grammar identified in the
caption in Old.

\section{Multiple alignments and information compression}\label{MA_compress}

This section describes in broad terms how alignments are evaluated
in terms of IC. A more detailed account of the method of evaluation used
in the SP61 model is given in Appendix \ref{appendix_B}.

Although {\raggedright IC and related concepts of probability are
well-established in the evaluation of alignments in bio-informatics
(see, for example, \cite{r24, r14, r3, r9, r2, r39}), the
framework here is different (as described in Section
\ref{varieties_of_MA}) which means that existing methods cannot be
applied directly.

}

In the present work, a good alignment is one which allows an economical
coding of New in terms of the patterns in Old.  The compression method
exploits the elementary principle that a (relatively long) sequential
pattern which repeats two or more times in a body of information may be
replaced by a shorter identifier, `tag' or `code' associated
with that pattern in some kind of `dictionary' of patterns. In effect,
each instance of the pattern in the data is unified with the same
pattern as it appears in the repository of patterns. This is the basis
of all standard methods for IC (see \cite{r28}).

In the ICMAUS scheme, this principle can be applied at a single `level', as
in the majority of standard compression schemes, but it can also be applied
at an arbitrary number of `higher' levels. To see what this means, consider
the alignment shown in Figure \ref{basic1}.

At the most basic level, a word like `t h i s' in New (the sentence
being parsed) is matched by the pattern `D 0 t h i s \#D' in Old (the
grammar) which means that the symbols `D 0 \#D' can be used as a `code'
for the pattern.\footnote{Although the code `D 0 \#D' does not appear
to be much smaller than `t h i s' in New, a weighting factor ensures
that the number of bits to be encoded is significantly larger than the
code, as explained in Appendix \ref{appendix_B}.}

A certain amount of compression can be achieved by encoding the words
in the sentence being parsed at a single level. But more compression
can be achieved by taking advantage of the fact that the words in the
sentence are not a random sequence of words but they conform to
grammatical patterns defined in the grammar like `NP D \#D N \#N \#NP'
and `S NP \#NP V \#V NP \#NP \#S'. The details of how this may be done
are explained in Appendix \ref{appendix_B}. As indicated above, this
kind of encoding at a `higher' level can be applied through arbitrarily
many levels, depending on the patterns of redundancy in New and in the
language from which it comes.

\section{The SP61 model}\label{the_SP61_model}

Given the example sentence discussed earlier (shown at the top of
Figure \ref{basic1}) and the grammar in Figure \ref{grammar_2}, the
SP61 model can find the alignment shown in Figure \ref{basic1} and, in
terms of compression, it identifies it as the best alignment amongst
the several which it forms for the given sentence and the given
grammar. Given relevant sentences and grammars, the model finds all
the other alignments shown in this article (they are indeed taken
directly from the output of the model).  In each case, the alignments
shown are the best in terms of IC amongst alternative alignments that
the model finds for a given sentence and grammar.

It is interesting to see that, in general, alignments that are good in terms of IC
are also `correct' in terms of our linguistic intuitions. This
relationship holds true for several other examples of parsing by the
model. Space limitations prevents them being shown here but
they can be found in \cite{r34b}.

\subsection{How the model works}

The SP61 model works by building alignments in a pairwise fashion
selecting the `best' in terms of compression at each stage. The method
thus constitutes a fairly straightforward application of
`metrics-guided' search: examine large search spaces in stages,
narrowing the search progressively at each stage using some kind of
`search metric' to guide the search. This accords with the need for
constraints in searching what is normally an astronomically large space
of possible alignments (Section \ref{search_and_constraints}).

Alignments can be built up in a pairwise manner because, at every
stage, new alignments are accepted only if they can `project' into a
one-dimensional pattern as described in Section
\ref{parsing_as_alignment}. Since any such alignment can be treated as
a single sequence of symbols it is possible to match it against any of
the original patterns in the grammar or any of the alignments formed at
earlier stages.

The program starts by searching for `good' alignments between the
sentence to be parsed and patterns in the grammar. For the example in
Figure \ref{basic1}, the best alignments found at this stage are between the
individual words in the sentence and corresponding patterns in the
grammar.

At the next stage, the program looks for `good' alignments between the
best of the alignments previously found and patterns in the grammar.
The `best' alignments at this stage are ones between the alignments
corresponding to the words and `higher level' patterns in the grammar.
Thus `D 0 t h i s \#D' and `N 1 b o y \#N' form an alignment with `NP D
\#D N \#N \#NP', giving `NP D 0 t h i s \#D N 0 b o y \#D \#NP'; likewise,
`V 0 l o v e s \#V' forms an alignment with `S NP \#NP V \#V NP \#NP \#S'
giving `S NP \#NP V 0 l o v e s \#V NP \#NP \#S'; then `D 1 t h a t \#D' and
`N 0 g i r l \#N' form an alignment with `NP D \#D N \#N \#NP' giving `NP D
1 t h a t \#D N 0 g i r l \#N \#NP'.

Finally, `NP D 0 t h i s \#D N 1 b o y \#D \#NP' and `NP D 1 t h a t
\#D N 1 g i r l \#N \#NP' are aligned with `S NP \#NP V 0 l o v e s \#V
NP \#NP \#S' giving the result shown in Figure \ref{basic1}. At each
stage, many `worse' alignments are formed which are weeded out by the
selection process.

An outline of how the model works is shown as pseudocode in Appendix
\ref{appendix_C} together with explanatory text.

\subsection{Computational complexity}\label{computational_complexity}

Given the well-known computational demands of multiple alignment
problems, readers may reasonably ask whether the proposed framework for
parsing would scale up to handle realistically large grammars and
longer sentences.

Estimates of the time complexity and space complexity of the model are
given here largely without justification owing to shortage of space. In a
serial processing environment, the time complexity of the model has
been estimated \cite{r34c} to be approximately O$(log_2 n \times nm)$,
where $n$ is the length of the sentence (in bits) and $m$ is the sum of
the lengths of the patterns in the grammar (in bits). In a parallel
processing environment, the time complexity may approach O$(log_2 n
\times n)$, depending on how the parallel processing is applied. In
serial and parallel environments, the space complexity should be
O$(m)$.

These estimates are based on the assumption that any given sentence is
processed as a single entity. However, the program has been designed so
that it is possible to process any given sentence as a succession of
`windows' (see Appendix \ref{windows}). Since it is possible to discard
all but the best intermediate results at the end of each window, the
time complexity of the model in a serial environment and operating in
`windows' mode appears to be approximately O($nm$). The time complexity
of the program in `windows' mode in a parallel environment depends on
exactly how the parallelism is applied but, in general, it is likely to
be better than in a serial environment.

\section{Decoding by compression: the production of
language}\label{decoding_by_compression}

As described in Appendix \ref{appendix_B}, a succinct,
coded representation of a sentence may be derived from a `good'
alignment amongst a set of sequences which includes the sentence and
rules in an appropriate grammar. This section proposes an idea which at
first sight may seem contradictory or paradoxical: that the decoding of
a coded representation of a sentence may be achieved by precisely the
same process of compression (by multiple alignment, unification and
search) as was used to achieve the original encoding! Although this may
superficially appear to be nonsense, careful reading of this section
should convince readers that the proposal is sound and that no laws of
logic or mathematics have been violated.

In this reversal of the original process of encoding, a sentence may be
created by finding a `good' alignment amongst a set of patterns that
includes a pattern that encodes the sentence (in New) together with
rules in the grammar which were used to create the encoding (in Old).
In both cases (encoding and decoding), alignments may be evaluated in
terms of the potential compression of one sequence: the sentence in the
first case and the encoded representation of the sentence in the second
case.

Figure \ref{basic2} shows an alignment of this kind produced by the
SP61 model. At the top of the figure is the sequence `S 0 1 0 1 0 \#S'
which is the encoded version of `t h i s b o y l o v e s t h a t g i r
l', as described in Appendix \ref{appendix_B}. The other sequences in
the figure are rules from the grammar shown in Figure \ref{grammar_2}.

\begin{figure}[b!hpt]
\centering
\begin{BVerbatim}
0 S      0              1                0              0
  |      |              |                |               
1 S NP   |              |          #NP V |           #V 1
    |    |              |           |  | |           |   
2   |    |              |           |  V 0 l o v e s #V 2
    |    |              |           |                    
3   |    |              |           |                   3
    |    |              |           |                    
4   |    |              |           |                   4
    |    |              |           |                    
5   |    |              |           |                   5
    |    |              |           |                    
6   |  D 0 t h i s #D   |           |                   6
    |  |           |    |           |                    
7   NP D           #D N |       #N #NP                  7
                      | |       |                        
8                     N 1 b o y #N                      8
                                                                
0      1              0                #S 0
       |              |                |   
1 NP   |              |            #NP #S 1
  |    |              |             |      
2 |    |              |             |     2
  |    |              |             |      
3 |    |            N 0 g i r l #N  |     3
  |    |            |           |   |      
4 NP D |         #D N           #N #NP    4
     | |         |                         
5    D 1 t h a t #D                       5
                                           
6                                         6
                                           
7                                         7
                                           
8                                         8
\end{BVerbatim}
\caption{\small The best alignment found by SP61 with `S 0 1 0 1 0 \#S' in New
and the grammar from Figure \ref{grammar_2} in Old.}
\label{basic2}
\normalsize
\end{figure}

As with parsing (Section \ref{parsing_as_multiple_alignment}), an
alignment may be interpreted by projecting its constituent symbols into
a single sequence. In the case of the alignment in Figure \ref{basic2},
the result of this projection is exactly the same as was shown in
Section \ref{parsing_as_alignment}. Although this sequence contains
grammatical symbols other than words, it has the right words in the
right order and may thus be regarded as a realisation of the sentence
corresponding to the coded sequence `S 0 1 0 1 0 \#S'.

\subsection{Decompression by compression}\label{decompression_by_compression}

The alignment shown in Figure \ref{basic2} achieves the paradoxical
effect of `decompression by compression' because the `input' (in New)
is a compressed code for a sentence and the `output' is an alignment
whose unification contains the original uncompressed sentence (together
with `service' symbols like `S', `NP' etc).

How can this paradox be resolved and how is it possible to achieve compression
with something (the code for the sentence) which is already compressed?

This is not as mysterious as it may at first sight seem.  The answer to
the riddle is the provision of two distinct sizes for each
symbol, as described in Appendices \ref{individual_symbols} and
\ref{summary_of_method}. The {\it minimum} size (in bits) is the
theoretical minimum calculated according to the S-F-E method, while the
{\it actual} size (in bits), which is the real size of the symbol in a
practical system, is larger than the minimum size by some constant
factor.

In the calculation of the compression difference (CD) for each alignment
(described in Appendix \ref{summary_of_method}), the actual sizes of
symbols are used to compute $B_N$, the number of bits required to
represent, in `raw' form, the symbols from New that enter into the
alignment. But the minimum sizes of symbols are used to compute $B_E$,
the number of bits required to encode the alignment. Thus the CD which
is derived from $B_N$ and $B_E$ represents the maximum compression
which is theoretically possible (with the given alignment within the
ICMAUS framework).

Given the distinction between a theoretical minimum size for each
symbol and a larger actual size, and given this way of calculating CD,
the alignment method that was used for the original parsing can be
used, without any modification, to find the best alignment for the code
for the sentence (in terms of CD values) and to discriminate it from
the many `wrong' alignments that are possible.

The foregoing remarks reflect what appears to be a general truth about
IC:  if lossless compression of a body of information is required (so
that the original form of the information can be reconstituted) then it
seems that the encoded form of the information must always contain some
residual redundancy. The existence of this residual redundancy may not
always be obvious but it seems that decoding is not possible without
it.

\section{Context sensitive aspects of syntax}\label{context_sensitive_aspects}

The examples considered so far may have given the impression that the
ICMAUS framework is merely a trivial variation on CF-PSG. This section
presents alignments from two areas of syntax showing how the ICMAUS
framework as it is realised in the SP61 model may accommodate `context
sensitive' aspects of syntax.

\subsection{Syntactic dependencies in French}

It often happens in natural languages that there are syntactic
dependencies between one part of a sentence and another. For example,
there is usually a `number' dependency between the subject of a
sentence and the main verb of the sentence: if the subject has a {\it
singular} form then the main verb must have a singular form and
likewise for {\it plural} forms of subject and main verb.

A prominent feature of these kinds of dependency is that they are often
`discontinuous' in the sense that the elements of the depency can be
separated, one from the next, by arbitrarily large amounts of
intervening structure. For example, the subject and main verb of a
sentence must have the same number (singular or plural) regardless of
the size of qualifying phrases or subortinate clauses that may come
between them.

Another interesting feature of syntactic dependencies is that one kind
of dependency (e.g., number dependency) can overlap other kinds of
dependency (e.g., gender ({\it masculine}/{\it feminine}) dependency),
as can be seen in the following example.

In the French sentence {\it Les plumes sont vertes} (``The feathers are
green'') there are two sets of overlapping syntactic dependencies like
this:

\begin{center}
\begin{BVerbatim}
 P        P  P          P        Number dependencies
Les plume s sont vert e s
      F               F          Gender dependencies
\end{BVerbatim}
\end{center}

\noindent In this example, there is a number dependency, which is
plural (`P') in this case, between the subject of the sentence, the
main verb and the following adjective: the subject is expressed with a
plural determiner ({\it Les}) and a noun ({\it plume}) which is marked
as plural with the suffix ({\it s}); the main verb ({\it sont}) has a
plural form and the following adjective ({\it vert}) is marked as
plural by the suffix ({\it s}).  Cutting right across these number
dependencies is the gender dependency, which is feminine (`F') in this
case, between the feminine noun ({\it plume}) and the adjective ({\it
vert}) which has a feminine suffix ({\it e}).

{\sloppy For many years, linguists puzzled how these kinds of syntactic
dependency could be represented succinctly in grammars for natural
languages. But then elegant solutions were found in Transformational
Grammar (TG, \cite{r10a}) and, later, in systems like Definite Clause
Grammars (DCG, \cite{r23}), based on Prolog.}

The solution proposed here is different from any established system and
is arguably simpler and more transparent than other systems. It will be
described and illustrated with a fragment of the grammar of French
which can generate the example sentence just shown. This fragment of
French grammar, shown in Figure \ref{fr10_grammar}, is expressed with
`patterns' in the same manner as the grammar in Figure \ref{grammar_2}
and others in this article.

\begin{figure}[b!hpt]
\centering
\begin{BVerbatim}
S NP #NP VP #VP #S (500)
NP D #D N #N #NP (700)
VP 0 V #V A #A #VP (300)
VP 1 V #V P #P NP #NP #VP (200)
P 0 sur #P (50)
P 1 sous #P (150)
V SNG est #V (250)
V PL sont #V (250)
D SNG M 0 le #D (90)
D SNG M 1 un #D (120)
D SNG F 0 la #D (130)
D SNG F 1 une #D (110)
D PL 0 les #D (125)
D PL 1 des #D (125)
N NR #NR NS1 #NS1 #N (450)
NS1 SNG - #NS1 (250)
NS1 PL s #NS1 (200)
NR M papier #NR (300)
NR F plume #NR (400)
A A AR #AR AS1 #AS1 AS2 #AS2 #A (300)
AS1 F e #AS1 (100)
AS1 M - #AS1 (200)
AS2 SNG - #AS2 (175)
AS2 PL s #AS2 (125)
AR 0 noir #AR (100)
AR 1 vert #AR (200)
NP SNG SNG #NP (450)
NP PL PL #NP (250)
NP M M #NP (450)
NP F F #NP (250)
N SNG V SNG A SNG (250)
N PL V PL A PL (250)
N M V A M (300)
N F V A F (400)
\end{BVerbatim}
\caption{\small A fragment of French grammar with patterns
for number dependencies and gender dependencies.}
\label{fr10_grammar}
\normalsize
\end{figure}

Apart from the use of patterns as the medium of expression, this
grammar differs from systems like TG or DCGs because the parts of the
grammar which express the forms of `high level' structures like
sentences, noun phrases and verb phrases (represented by the first four
patterns in Figure \ref{fr10_grammar}) do not contain any reference to
number or gender.

Instead, the grammar contains patterns like `NP SNG SNG \#NP' and `N M
V A M' (the last eight patterns in Figure \ref{fr10_grammar}). The
first of these says, in effect, that between the symbols `NP' and
`\#NP' there are two structures marked as singular (`SNG'). In this
simple grammar, there is no ambiguity about what those two structures
are:  they can only be a determiner (`D') followed by a noun (`N'). In
a more complex grammar, there would need to be disambiguating
context to establish the `correct' alignments of symbols.  The second
pattern says, in effect, that in a sentence which contains the
(discontinuous) sequence of symbols `N V A', the noun (`N') is
masculine (`M') and the adjective (`A') is also masculine.

\subsubsection{An alignment}

\begin{figure}[b!hpt]
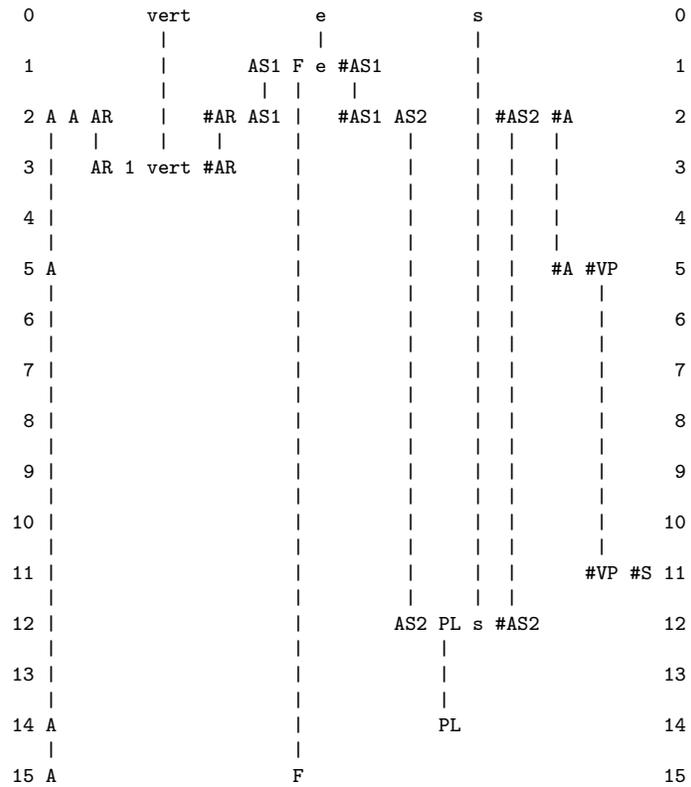

\fontsize{08.00pt}{09.60pt}
\centering
\begin{BVerbatim}
 0          vert           e             s                 0
             |             |             |                  
 1           |       AS1 F e #AS1        |                 1
             |        |  |    |          |                  
 2 A A AR    |   #AR AS1 |   #AS1 AS2    | #AS2 #A         2
   |   |     |    |      |         |     |  |   |           
 3 |   AR 1 vert #AR     |         |     |  |   |          3
   |                     |         |     |  |   |           
 4 |                     |         |     |  |   |          4
   |                     |         |     |  |   |           
 5 A                     |         |     |  |   #A #VP     5
   |                     |         |     |  |       |       
 6 |                     |         |     |  |       |      6
   |                     |         |     |  |       |       
 7 |                     |         |     |  |       |      7
   |                     |         |     |  |       |       
 8 |                     |         |     |  |       |      8
   |                     |         |     |  |       |       
 9 |                     |         |     |  |       |      9
   |                     |         |     |  |       |       
10 |                     |         |     |  |       |     10
   |                     |         |     |  |       |       
11 |                     |         |     |  |      #VP #S 11
   |                     |         |     |  |               
12 |                     |        AS2 PL s #AS2           12
   |                     |            |                     
13 |                     |            |                   13
   |                     |            |                     
14 A                     |            PL                  14
   |                     |                                  
15 A                     F                                15
\end{BVerbatim}
\caption{\small The best alignment found by SP61 with `les plume s sont
vert e s' in New and the grammar from Figure \ref{fr10_grammar}
in Old (Part 2).}
\label{alignment_fr10_2}
\normalsize
\end{figure}

The alignment in Figures \ref{alignment_fr10_1} and
\ref{alignment_fr10_2} shows the best alignment found by SP61 with our
example sentence in New and the grammar from Figure \ref{fr10_grammar}
in Old.\footnote{By contrast with the alignments shown in Figures
\ref{basic1} and \ref{basic2}, the alignment in Figures
\ref{alignment_fr10_1} and \ref{alignment_fr10_2}, and all subsequent
alignments in this article, were originally created with spaces between
the letters in every word, as in Figures \ref{basic1} and
\ref{basic2}.  However, for the sake of readability (as suggested by
one of the referees) and to save space, the alignments have been
prepared again with no spaces within words (except where suffixes need
to be identified as distinct entities within the grammar).} The main
constituents of the sentence are marked in an appropriate manner and
dependencies for number and gender are marked by patterns appearing in
rows 13, 14 and 15 of the alignment.

\begin{figure}[b!hpt]
\fontsize{08.00pt}{09.60pt}
\centering
\begin{BVerbatim}
 0             les           plume            s                       sont     0
                |              |              |                        |        
 1              |              |              |                        |       1
                |              |              |                        |        
 2              |              |              |                        |       2
                |              |              |                        |        
 3              |              |              |                        |       3
                |              |              |                        |        
 4              |              |              |                  V PL sont #V  4
                |              |              |                  | |       |    
 5              |              |              |             VP 0 V |       #V  5
                |              |              |             |    | |            
 6              |              |       NS1 PL s #NS1        |    | |           6
                |              |        |  |     |          |    | |            
 7              |     N NR     |   #NR NS1 |    #NS1 #N     |    | |           7
                |     | |      |    |      |         |      |    | |            
 8              |     | NR F plume #NR     |         |      |    | |           8
                |     |    |               |         |      |    | |            
 9      D PL 0 les #D |    |               |         |      |    | |           9
        | |        |  |    |               |         |      |    | |            
10   NP D |        #D N    |               |         #N #NP |    | |          10
     |    |           |    |               |             |  |    | |            
11 S NP   |           |    |               |            #NP VP   | |          11
     |    |           |    |               |             |       | |            
12   |    |           |    |               |             |       | |          12
     |    |           |    |               |             |       | |            
13   NP   PL          |    |               PL           #NP      | |          13
                      |    |               |                     | |            
14                    N    |               PL                    V PL         14
                      |    |                                     |              
15                    N    F                                     V            15
\end{BVerbatim}
\caption{\small The best alignment found by SP61 with `les plume s sont
vert e s' in New and the grammar from Figure \ref{fr10_grammar}
in Old (Part 1).}
\label{alignment_fr10_1}
\normalsize
\end{figure}

\subsubsection{Discussion}

Readers may wonder why, in the example just shown, the pattern `NP PL
PL \#NP' is separate from the pattern `N PL V PL A PL'. Why not simply
merge them into something like `NP PL N PL \#NP V PL A PL'. The reason
for separating the number dependencies in noun phrases (`NP') from the
other number dependencies is that they do no always occur together. For
example, noun phrases may be found within one of the two verb-phrase
(`VP') patterns shown in Figure \ref{fr10_grammar} (the fourth pattern
in the grammar) and this context does not contain the `N ... V ... A
...' pattern.

Another question that may come to mind is what happens when there are
one or more subordinate clauses between the subject of a sentence and
the main verb of the sentence, and when there are verbs in the
subordinate clauses. In the case of number dependencies between subject
and main verb, how can the system distinguish between the main verb and
one of the verbs in the subordinate clauses? There is insufficient
space here for a full answer to this question. In brief, it seems that this
kind of problem can be overcome by providing disambiguating context in
the patterns that express number dependency (see \cite{r34b}).

These ideas are still relatively new and there is plenty of scope for
further investigation and development.

\subsection{Dependencies in the syntax of English auxiliary verbs}

This subsection presents a grammar and examples showing how the syntax of
English auxiliary verbs may be described in the ICMAUS framework.
Before the grammar and examples are presented, the syntax of this part
of English is described and alternative formalisms for describing the
syntax are briefly discussed.

In English, the syntax for main verbs and the `auxiliary' verbs which
may accompany them follows two quasi-independent patterns of constraint
which interact in an interesting way.

The {\it primary pattern of constraint} may be expressed with this sequence of symbols,

\begin{center}
M H B B V,
\end{center}

\noindent which should be interpreted in the following way:

\begin{itemize}

\item Each letter represents a category for a single word:

\begin{itemize}

\item `M' stands for `modal' verbs like `will', `can', `would' etc.

\item `H' stands for one of the various forms of the verb `to have'.

\item Each of the two instances of `B' stands for one of the various
forms of the verb `to be'.

\item `V' stands for the main verb which can be any verb except a modal
verb (except, arguably, when it occurs by itself).

\end{itemize}

\item The words occur in the order shown but any of the words may be
omitted.

\item \sloppy Questions of `standard' form follow exactly the same pattern as
statements except that the first verb, whatever it happens to be (`M',
`H', the first `B', the second `B' or `V'), precedes the subject noun
phrase instead of following it.

\end{itemize}

Here are two examples of the primary pattern with all of the words
included:

\begin{center}
\begin{BVerbatim}
It will have been being washed  
    M    H    H     B     V   
 
Will it have been being washed?  
 M       H    H     B     V   
\end{BVerbatim}
\end{center}

The {\it secondary constraints} are these:

\begin{itemize}

\item Apart from the modals, which always have the same form, the first
verb in the sequence, whatever it happens to be (`H', the first `B',
the second `B' or `V'), always has a `finite' form (the form it would
take if it were used by itself with the subject).

\item If an `M' auxiliary verb is chosen, then whatever follows it
(`H', first `B', second `B', or `V') must have an `infinitive' form
(i.e., the `standard' form of the verb as it occurs in the context `to
...', but without the word `to').

\item If an `H' auxiliary verb is chosen, then whatever follows it (the
first `B', the second `B' or `V') must have a past tense form such as
`been', `seen', `gone', `slept', `wanted' etc. In Chomsky's {\it
Syntactic Structures} \cite{r10a}, these forms were characterised as
{\it en} forms and the same convention has been adopted here.

\item If the first of the two `B' auxiliary verbs is chosen, then
whatever follows it (the second `B' or `V') must have an {\it ing} form,
e.g., `singing', `eating', `having', `being' etc.

\item If the second of the two `B' auxiliary verbs is chosen, then
whatever follows it (only the main verb is possible now) must have a
past tense form (marked with {\it en} as above).

\item The constraints apply to questions in exactly the same way as they
do to statements.

\end{itemize}

Figure \ref{english_sentences} shows a selection of examples with the
dependencies marked.

\begin{figure}[b!hpt]
\centering
\begin{BVerbatim}
           H------en  B2---------en
          ----    --  --         --
It  will  have  been  being  washed
    ----  ----  --      ---  ----
     M----inf   B1------ing   V


          B1------ing
          --      ---
Will  he  be  talking?
----      --  ----
 M-------inf   V


              V
            ------
They  have  finished
      ----        --
       H----------en
      fin


Are  they  gone?
---        ----
B2----------en
fin         V


         B1--------ing
         --        ---
Has  he  been  working?
---        --  ----
 H---------en   V
fin
\end{BVerbatim}
\caption{\small A selection of example sentences in English with
markings of dependencies between the verbs. {\it Key:} {\it M} = modal,
{\it H} = forms of the verb `have', {\it B1} = first instance of a form
of the verb `be', {\it B2} = second instance of a form of the verb
`be', {\it V} = main verb, {\it fin} = a finite form, {\it inf} = an
infinitive form, {\it en} = a past tense form, {\it ing} = a verb
ending in `ing'.}
\label{english_sentences}
\normalsize
\end{figure}

\subsubsection{Transformational grammar and English auxiliary verbs}

In Figure \ref{english_sentences} it can be seen that in many cases but
not all, the dependencies which have been described may be regarded as
discontinuous because they connect one word in the sequence to the
suffix of the following word thus bridging the stem of the following
word. Three instances of this discontinuous kind of dependency can be
seen in the first example in the figure.

In {\it Syntactic Structures}, \cite{r10a} showed that this kind of
regularity in the syntax of English auxiliary verbs could be described
using Transformational Grammar (TG). For each pair of symbols linked by
a dependency (`M inf', `H en', `B1 ing', `B2 en') the two symbols could
be shown together in the `deep structure' of a sentence and then moved
into their proper position or modified in form (or both) using
`transformational rules'.

This elegant demonstration argued persuasively in favour of TG compared
with alternatives which were available at that time. However, later
research has shown that the same kinds of regularities in the syntax of
English auxiliary verbs can be described quite well without recourse to
transformational rules, using Definite Clause Grammars (DCGs) or other
systems which do not use that type of rule (see, for example,
\cite{r23, r17}). An example showing how English auxiliary verbs
may be described using the DCG formalism may be found in
\cite[pp. 183-184]{r45a}).

\subsubsection{English auxiliary verbs in the ICMAUS framework}

Figures \ref{auxiliary_verbs_1} and \ref{auxiliary_verbs_2} show an
`ICMAUS' grammar for English auxiliary verbs which exploits several of
the ideas described earlier in this article. Figure
\ref{three_parsings_1}, Figures \ref{three_parsings_2_a} and
\ref{three_parsings_2_b}, and Figure \ref{three_parsings_3} show the
best alignments in terms of IC for three different sentences produced
by the SP61 model using this grammar. In the following paragraphs,
aspects of the grammar and of the examples are described and
discussed.

\begin{figure}[b!hpt]
\centering
\begin{BVerbatim}
S ST NP #NP X1 #X1 XR #S (3000)
S Q X1 #X1 NP #NP XR #S (2000)
NP SNG it #NP (4000)
NP PL they #NP (1000)
X1 0 V M #V #X1 XR XH XB XB XV #S (1000)
X1 1 XH FIN #XH #X1 XR XB XB XV #S (900)
X1 2 XB1 FIN #XB1 #X1 XR XB XV #S (1900)
X1 3 V FIN #V #X1 XR #S (900)
XH V H #V #XH XB #S (200)
XB XB1 #XB1 XB #S (300)
XB XB1 #XB1 XV #S (300)
XB1 V B #V #XB1 (500)
XV V #V #S (5000)
M INF (2000)
H EN (2400)
B XB ING (2000)
B XV EN (700)
SNG SNG (2500)
PL PL (2500)
\end{BVerbatim}
\caption{\small A grammar for the syntax of English
auxiliary verbs (Part 1).}
\label{auxiliary_verbs_1}
\normalsize
\end{figure}

\begin{figure}[b!hpt]
\centering
\begin{BVerbatim}
V M 0 will #V (2500)
V M 1 would #V (1000)
V M 2 could #V (500)
V H INF have #V (600)
V H PL FIN have #V (400)
V H SNG FIN has #V (200)
V H EN had #V (500)
V H FIN had #V (300)
V H ING hav ING1 #ING1 #V (400)
V B SNG FIN 0 is #V (500)
V B SNG FIN 1 was #V (400)
V B INF be #V (400)
V B EN be EN1 #EN1 #V (600)
V B ING be ING1 #ING1 #V (700)
V B PL FIN 0 are #V (300)
V B PL FIN 1 were #V (500)
V FIN wrote #V (166)
V INF 0 write #V (254)
V INF 1 chew #V (138)
V INF 2 walk #V (318)
V INF 3 wash #V (99)
V ING 0 chew ING1 #ING1 #V (623)
V ING 1 walk ING1 #ING1 #V (58)
V ING 2 wash ING1 #ING1 #V (102)
V EN 0 made #V (155)
V EN 1 brok EN1 #EN1 #V (254)
V EN 2 tak EN1 #EN1 #V (326)
V EN 3 lash ED #ED #V (160)
V EN 4 clasp ED #ED #V (635)
V EN 5 wash ED #ED #V (23)
ING1 ing #ING1 (1883)
EN1 en #EN1 (1180)
ED ed #ED (818)
\end{BVerbatim}
\caption{\small A grammar for the syntax of English
auxiliary verbs (Part 2).}
\label{auxiliary_verbs_2}
\normalsize
\end{figure}

\subsubsection{The primary constraints}

\begin{figure}[b!hpt]
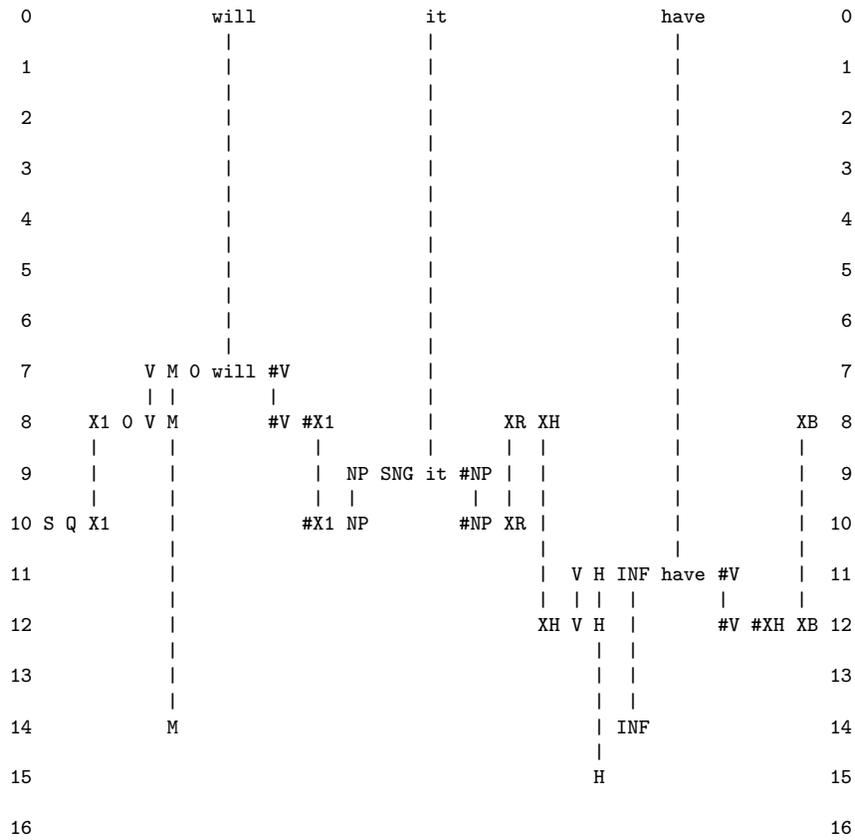

\fontsize{08.00pt}{09.60pt}
\centering
\begin{BVerbatim}
 0                will               it                   have            0
                   |                 |                     |               
 1                 |                 |                     |              1
                   |                 |                     |               
 2                 |                 |                     |              2
                   |                 |                     |               
 3                 |                 |                     |              3
                   |                 |                     |               
 4                 |                 |                     |              4
                   |                 |                     |               
 5                 |                 |                     |              5
                   |                 |                     |               
 6                 |                 |                     |              6
                   |                 |                     |               
 7          V M 0 will #V            |                     |              7
            | |        |             |                     |               
 8     X1 0 V M        #V #X1        |      XR XH          |          XB  8
       |      |            |         |      |  |           |          |    
 9     |      |            |  NP SNG it #NP |  |           |          |   9
       |      |            |  |          |  |  |           |          |    
10 S Q X1     |           #X1 NP        #NP XR |           |          |  10
              |                                |           |          |    
11            |                                |  V H INF have #V     |  11
              |                                |  | |  |       |      |    
12            |                                XH V H  |       #V #XH XB 12
              |                                     |  |                   
13            |                                     |  |                 13
              |                                     |  |                   
14            M                                     | INF                14
                                                    |                      
15                                                  H                    15
                                                                     
16                                                                       16
\end{BVerbatim}
\caption{\small The best alignment found by SP61 with `will it have
be en brok en' in New and the grammar from Figures \ref{auxiliary_verbs_1} and
\ref{auxiliary_verbs_2} in Old (Part 1).}
\label{three_parsings_2_a}
\normalsize
\end{figure}

\begin{figure}[b!hpt]
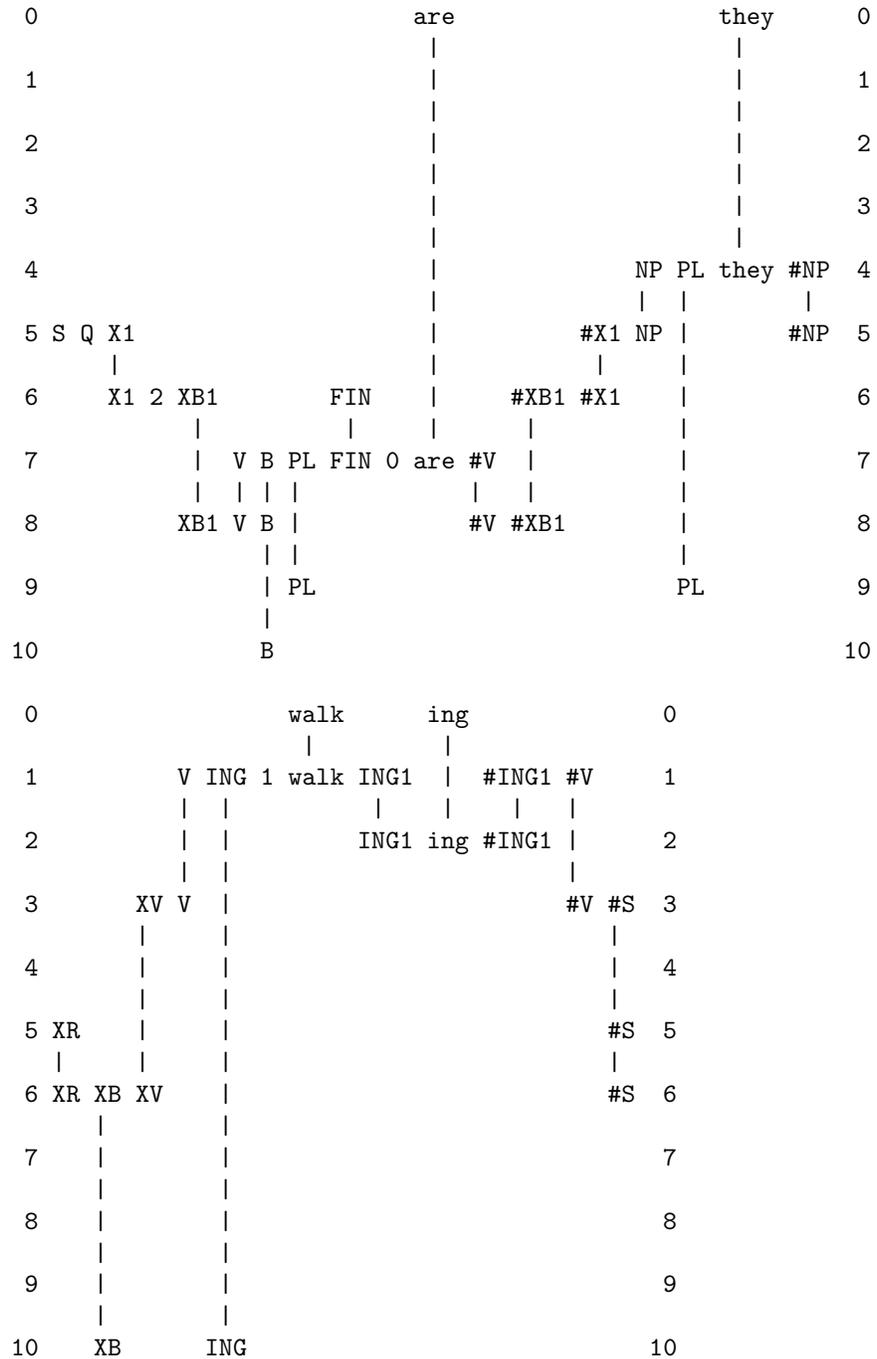

\centering
\begin{BVerbatim}
 0                           are                   they      0
                              |                     |         
 1                            |                     |        1
                              |                     |         
 2                            |                     |        2
                              |                     |         
 3                            |                     |        3
                              |                     |         
 4                            |              NP PL they #NP  4
                              |              |  |        |    
 5 S Q X1                     |          #X1 NP |       #NP  5
       |                      |           |     |             
 6     X1 2 XB1        FIN    |     #XB1 #X1    |            6
             |          |     |      |          |             
 7           |  V B PL FIN 0 are #V  |          |            7
             |  | | |            |   |          |             
 8          XB1 V B |            #V #XB1        |            8
                  | |                           |             
 9                | PL                          PL           9
                  |                                           
10                B                                         10

 0                  walk      ing              0
                     |         |                
 1          V ING 1 walk ING1  |  #ING1 #V     1
            |  |          |    |    |   |       
 2          |  |         ING1 ing #ING1 |      2
            |  |                        |       
 3       XV V  |                        #V #S  3
         |     |                           |    
 4       |     |                           |   4
         |     |                           |    
 5 XR    |     |                           #S  5
   |     |     |                           |    
 6 XR XB XV    |                           #S  6
      |        |                                
 7    |        |                               7
      |        |                                
 8    |        |                               8
      |        |                                
 9    |        |                               9
      |        |                                
10    XB      ING                             10
\end{BVerbatim}
\caption{\small The best alignment found by SP61 with `are they
walk ing' in New and the grammar from Figures \ref{auxiliary_verbs_1} and
\ref{auxiliary_verbs_2} in Old.}
\label{three_parsings_3}
\normalsize
\end{figure}

The first line in the grammar is a sentence pattern for a statement
(marked with the symbol `ST') and the second line is a sentence pattern
for a question (marked with the symbol `Q'). Apart from these markers,
the only difference between the two patterns is that, in the statement
pattern, the symbols `X1 \#X1' follow the noun phrase symbols (`NP
\#NP'), whereas in the question pattern they precede the noun phrase
symbols. As can be seen in the examples in Figure
\ref{three_parsings_1}, Figures \ref{three_parsings_2_a} and
\ref{three_parsings_2_b}, and Figure \ref{three_parsings_3}, the pair of
symbols, `X1 \#X1', has the effect of selecting the first verb in the
sequence of auxiliary verbs and ensuring its correct position with
respect to the noun phrase. In Figure \ref{three_parsings_1} it follows
the noun phrase, while in Figures \ref{three_parsings_2_a} and
\ref{three_parsings_2_b}, and Figure \ref{three_parsings_3} it precedes
the noun phrase.

\begin{figure}[b!hpt]
\centering
\begin{BVerbatim}
 0             it                            is                    0
               |                             |                      
 1             |                             |                     1
               |                             |                      
 2             |                             |                     2
               |                             |                      
 3             |                             |                     3
               |                             |                      
 4             |                 B           |                     4
               |                 |           |                      
 5             |               V B SNG FIN 0 is #V                 5
               |               | |  |   |       |                   
 6             |           XB1 V B  |   |       #V #XB1            6
               |            |       |   |           |               
 7             |      X1 2 XB1      |  FIN         #XB1 #X1 XR XB  7
               |      |             |                    |  |       
 8      NP SNG it #NP |             |                    |  |      8
        |   |      |  |             |                    |  |       
 9 S ST NP  |     #NP X1            |                   #X1 XR     9
            |                       |                               
10         SNG                     SNG                            10
                                                                      
 0           wash    ed            0
              |      |              
 1    V EN 5 wash ED |  #ED #V     1
      | |         |  |   |  |       
 2    | |         ED ed #ED |      2
      | |                   |       
 3 XV V |                   #V #S  3
   |    |                      |    
 4 XV   EN                     |   4
   |                           |    
 5 |                           |   5
   |                           |    
 6 |                           |   6
   |                           |    
 7 XV                          #S  7
                               |    
 8                             |   8
                               |    
 9                             #S  9
                                      
10                                10
\end{BVerbatim}
\caption{\small The best alignment found by SP61 with `it is wash ed' in
New and the grammar from Figures \ref{auxiliary_verbs_1} and \ref{auxiliary_verbs_2} in Old.}
\label{three_parsings_1}
\normalsize
\end{figure}

\begin{figure}[b!hpt]
\fontsize{08.00pt}{09.60pt}
\centering
\begin{BVerbatim}     
 0               be     en                        brok     en             0
                 |      |                          |       |               
 1               |  EN1 en #EN1                    |       |              1
                 |   |      |                      |       |               
 2        V B EN be EN1    #EN1 #V                 |       |              2
          | | |                 |                  |       |               
 3    XB1 V B |                 #V #XB1            |       |              3
       |    | |                     |              |       |               
 4     |    | |                     |      V EN 1 brok EN1 |  #EN1 #V     4
       |    | |                     |      | |          |  |   |   |       
 5     |    | |                     |   XV V |          |  |   |   #V #S  5
       |    | |                     |   |    |          |  |   |      |    
 6 XB XB1   | |                    #XB1 XV   |          |  |   |      #S  6
   |        | |                         |    |          |  |   |      |    
 7 |        | |                         |    |          |  |   |      |   7
   |        | |                         |    |          |  |   |      |    
 8 XB       | |                         XV   |          |  |   |      #S  8
            | |                         |    |          |  |   |      |    
 9          | |                         |    |          |  |   |      |   9
            | |                         |    |          |  |   |      |    
10          | |                         |    |          |  |   |      #S 10
            | |                         |    |          |  |   |      |    
11          | |                         |    |          |  |   |      |  11
            | |                         |    |          |  |   |      |    
12          | |                         |    |          |  |   |      #S 12
            | |                         |    |          |  |   |           
13          | |                         |    |         EN1 en #EN1       13
            | |                         |    |                             
14          | |                         |    |                           14
            | |                         |    |                             
15          | EN                        |    |                           15
            |                           |    |                             
16          B                           XV   EN                          16
\end{BVerbatim}
\caption{\small The best alignment found by SP61 with `will it have
be en brok en' in New and the grammar from Figures \ref{auxiliary_verbs_1} and
\ref{auxiliary_verbs_2} in Old (Part 2).}
\label{three_parsings_2_b}
\normalsize
\end{figure}

Each of the next four patterns in the grammar have the form `X1 ... \#X1
XR ... \#S'. The symbols `X1' and `\#X1' align with the same pair of
symbols in the sentence pattern. The symbols `XR ... \#S' encode the
remainder of the sequence of verbs.

The first `X1' pattern encodes verb sequences which start with a modal
verb (`M'), the second one is for verb sequences beginning with a
finite form of the verb `have' (`H'), the third is for sequences
beginning with either of the two `B' verbs in the primary sequence (see
below), and the last `X1' pattern is for sentences which contain a main
verb without any auxiliaries.

In the first of the `X1' patterns, the subsequence `XR ... \#S' encodes
the remainder of the sequence of auxiliary verbs using the symbols `XH
XB XB XV'. In a similar way, the subsequence `XR ... \#S' within each of
the other `X1' patterns encodes the verbs which follow the first verb
in the sequence.

Notice that the pattern `X1 2 XB1 FIN \#XB1 \#X1 XR XB XV \#S' can
encode sentences which start with the first `B' verb and also contains
the second `B' verb. And it also serves for any sentence which starts
with the first or the second `B' verb with the omission of the other
`B' verb. In the latter two cases, the `slot' between the symbols `XB'
and `XV' is left vacant. Figure \ref{three_parsings_1} illustrates the
case where the verb sequence starts with the first `B' verb with the
omission of the second `B' verb. Figure \ref{three_parsings_3}
illustrates the case where the verb sequence starts with the second `B'
verb (and the first `B' verb has been omitted).

\subsubsection{The secondary constraints}

The secondary constraints are represented using the patterns `M INF',
`H EN', `B XB ING' and `B XV EN'. Singular and plural dependencies are
marked in a similar way using the patterns `SNG SNG' and `PL PL'.

Examples appear in all three alignments in Figure
\ref{three_parsings_1}, Figures \ref{three_parsings_2_a} and
\ref{three_parsings_2_b}, and Figure \ref{three_parsings_3}. In every
case except one (row 4 in Figure \ref{three_parsings_1}), the patterns
representing secondary constraints appear in the bottom rows of the
alignment. These examples show how dependencies bridging arbitrarily
large amounts of structure, and dependencies that overlap each other,
can be represented with simplicity and transparency in the medium of
multiple alignments.

Notice, for example, how dependencies between the first and second verb
in a sequence of auxiliary verbs are expressed in the same way
regardless of whether the two verbs lie side by side (e.g., the
statement in Figure \ref{three_parsings_1}) or whether they are
separated from each other by the subject noun-phrase (e.g., the
question in Figures \ref{three_parsings_2_a} and
\ref{three_parsings_2_b} and the question in Figure
\ref{three_parsings_3}).  Notice, again, how the overlapping
dependencies in Figures \ref{three_parsings_2_a} and
\ref{three_parsings_2_b} and their independence from each other are
expressed with simplicity and clarity in the ICMAUS framework.

Readers may wonder why the two patterns representing dependencies
between a `B' verb and whatever follows it (`B XB ING' and `B XV EN')
contain three symbols rather than two. One reason is that, when two (or
more) patterns begin with the same symbol (or sequence of symbols), the
scoring method for evaluating alignments requires that the two patterns
can be distinguished from each other by one (or more) symbols in each
pattern which does not include the terminal symbol in each pattern. A
second reason is that the second symbol in each pattern helps to
determine whether the `B' at the start of the pattern corresponds to
the first or the second `B' verb in the primary sequence:

\begin{itemize}

\item `B XB ING'. The inclusion of `XB' in this pattern means that the
`B' verb is the first of the two `B' verbs in the primary sequence and
the following verb must be `ING'.

\item `B XV EN'. The inclusion of `XV' in this pattern means that the
`B' verb may be the first or the second of the two `B' verbs. However,
since the first case is already covered by 'B XB ING', this pattern
covers the constraint between the second `B' verb and verbs of the
category `EN'.

\end{itemize}

\section{Discussion and Conclusion}\label{discussion_conclusion}

This section considers briefly a selection of topics relating to
the development of these ideas.

\subsection{Other examples}

In the space available, it has not been possible to show more than a
small selection of examples. Additional example may be found in
\cite{r34a, r34b} showing: how the system can find alternative parsings
when there are ambiguities in the text being parsed; how recursive
structures in syntax can be parsed; how the provision of appropriate
context can resolve ambiguities when discontinuous dependencies of one
type are nested, one within another; and one possible way in which
`cross-serial dependencies' in syntax may be represented and parsed in
the ICMAUS framework.

Other examples showing how the system can handle recursive structures
in syntax may be found in \cite{r33}.

\subsection{Parsing and learning}\label{parsing_and_learning}

As was noted in Section \ref{background_and_context}, much of the
thinking in this research programme is based on an earlier programme of
research into the unsupervised learning of linguistic structures
\cite{r43, r45, r46}. The ICMAUS framework and the SP61 model have been
developed with the express intention that they should accommodate
inductive learning and integrate it in a seamless manner with other
capabilities of the model.

It is envisaged that the framework will be developed so that, when New
information is received that cannot be unified fully with patterns in
Old, the patterns or parts of patterns in New which do not unify with
existing patterns in Old will be simply added to Old with
system-generated code symbols where appropriate. By hypothesis in this
research programme, the process of adding New knowledge to Old in a
manner which minimises redundancy (as far as is practically possible)
will capture the essentials of unsupervised inductive learning.

Development of the ICMAUS framework to incorporate inductive learning in this
way is currently in progress.

\subsection{Potential advantages of
using patterns to represent knowledge}\label{advantages_of_patterns}

In this research, `patterns' have been adopted as the medium for
representing knowledge:

\begin{itemize}

\item Because they seem to offer a good prospect of providing a
`universal' medium for representing diverse kinds of knowledge.

\item For similar reasons, `patterns' seem to lend themselves to the
representation of knowledge at both `concrete' and `abstract' levels of
abstraction.

\item For these reasons, the use of patterns may facilitate the
seamless integration of diverse kinds of knowledge over a wide range of
abstractions.

\item Likewise, the use of patterns may facilitate the development of a
learning system that can operate freely with diverse kinds of knowledge
over a wide range of abstractions.

\end{itemize}

\subsection{Integration and generalisation}

If, as suggested in Section \ref{background_and_context}, both linguistic and
non-linguistic structures may be accommodated naturally within the
ICMAUS framework, then grammars of the kind shown previously may, at
some stage, be extended seamlessly to include the `meanings' of
syntactic forms. Parsing and production of language as described here
should generalise without radical reorganisation to a more rounded
model of language understanding and production of language which
includes meanings.

In a similar way, the potential of the system noted in Section
\ref{background_and_context} to accommodate other aspects of
`intelligence' such as probabilistic and other kinds of reasoning,
best-match pattern recognition and inductive learning suggests
potential in the system for the eventual integration of natural
language processing with non-linguistic `intelligence' of various
kinds.

\subsection{Conclusion}

In this article I have tried to show informally with examples how the
representation of natural language syntax and the parsing and
production of natural language may be understood as ICMAUS.

A novel feature of these proposals is the superficially paradoxical
idea that a single process of information compression by multiple
alignment, unification and search may achieve both the encoding and the
decoding of information, both the analysis and the production of
sentences. This is not simply a gimmick: in practical terms it offers
the prospect that one search engine may be used for both purposes and
it offers a theoretical bonus in extending the explanatory range of the
model without the need for any {\it ad hoc} additions or
modifications.

The suggested method of representing the syntax of natural language
appears to be simpler and more direct than existing methods. This
method may have benefits in the creation of hand-crafted grammars for
natural languages. Perhaps more significantly, it may simplify the
automatic learning of grammars for natural languages which is envisaged
in the further development of these ideas.

In general, an important motivation for further development of these
ideas is the potential which they offer for the integration of parsing
and production of language with other aspects of computing including
unsupervised learning, deductive and probabilistic inference, (fuzzy)
pattern recognition, (fuzzy) information retrieval and others. In the
broadest terms, the aim of this research programme and a touchstone for
its success or failure is the development of a model which exhibits a
favourable combination of conceptual {\it simplicity} with explanatory
or descriptive {\it power}.

\section*{Acknowledgements}

I am grateful to Prof. C. S Wallace of Monash University for discussion
of some of the ideas presented in this article, to Dr Chris Mellish of the
Deparment of Artificial Intelligence (Division of Informatics), University
of Edingburgh, for useful comments and advice, and to Dr. Tim Porter and
Mr. John Hornsby both of the School of Informatics, University of
Wales at Bangor, for constructive comments on an earlier version of
this article. I am grateful also to James Crook of Dublin for positive
comments and constructive suggestions about these ideas.

%
%

\appendix

\section{Definitions of terms}\label{appendix_A}

\subsection{`Symbol'}

A {\it symbol} is some kind of mark which can be compared with any other symbol. In the context of pattern matching, a symbol is the smallest unit which can participate in matching: a symbol can be compared (matched) only with another single symbol and the result of matching is either that the two symbols are the same or that they are different. No other result is permitted.

An important feature of the concept of a {\it symbol}, as it is used in
this research, is that, with one qualification, it has {\it
no hidden meaning}. In this research, a symbol is a primitive mark
which can be discriminated in a yes/no manner from other symbols. There
are no symbols like the symbols in an arithmetic function (e.g., `6',
`22', `+', `-', `$\times$', `/', `(`, `)' etc), each of which has a
meaning for the user which is not directly visible.

The one qualification to the slogan ``no hidden meaning'' is that it seems necessary to allow the system to make a distinction, relative to each pattern, between symbols that are `code' for that pattern and symbols that are `data' or `contents' for the pattern. Labels like `code' or `data' reflect operations of the system itself (or some comparable system in the past) and may therefore be regarded as distinct from `user-oriented' meanings that are intrinsic to the material being processed.

For any given symbol (or group of symbols), it is possible to express
meanings of this latter kind but those meanings must take the form of
one or more additional symbols which are associated with the given
symbol (or group of symbols) and are thus explicit and visible within
the structure of symbols and patterns.

\subsubsection{`Symbol type' and `alphabet'}

If two symbols match, we say that they belong to the same {\it symbol type}. In any system which contains symbols, we normally recognise an {\it alphabet} of symbol types such that every symbol in the system belongs in one and only one of the symbol types in the alphabet, and every symbol type is represented at least once in the system.

\subsubsection{`Hit' and `gap'}

A positive match between two symbols is termed a {\it hit}. In any
given pattern in an alignment of two more {\it patterns}, one or more
unmatched symbols  between two hits in the pattern or before the first
or after the last hit is termed a {\it gap}.

\subsection{`Pattern'}

A {\it pattern} is an array of symbols in one, two or more dimensions. In this article, one dimensional patterns ({\it sequences} or {\it strings} of symbols) are the main focus of attention.

The meaning of the term {\it pattern} includes the meanings of the terms {\it substring} and {\it subsequence}, defined next.

\subsection{`Substring'}

A {\it substring} is a sequence of symbols of length $n$ within a sequence of length $m$, where $n \leqslant m$ and where the constituent symbols in the substring are contiguous within the sequence which contains the substring.

\subsection{`Subsequence'}

A {\it subsequence} is a sequence of symbols of length $n$ within a sequence of length $m$, where $n \leqslant m$ and where the constituent symbols in the subsequence may not be contiguous within the sequence which contains the subsequence. The set of all subsequences of a given sequence includes all the substrings of that sequence.

\subsection{`Alignment'}

In the case of one-dimensional patterns,\footnote{As previously noted,
the concept of an alignment may be generalised to patterns of two or
more dimensions. But no attempt is made here to provide a formal
definition for alignments of patterns of two dimensions or higher.} an
{\bf alignment} is a two-dimensional array of one or more sequences of
symbols, each one in a separate {\bf row} in the array. The alignment
shows sets of two or more matching symbols by arranging the symbols in
each set in a {\bf column} of the array.\footnote{The fact that, in
displaying alignments, it can sometimes be convenient to put
non-matching symbols in the same column with lines to mark the symbols
that do match (as in Figure \ref{DNA}) is not relevant to the abstract
definition of an alignment presented here.} In an alignment, as defined
in this research:

\begin{itemize}
\item Symbols which are contiguous in a pattern which appears in an alignment, need not occupy contiguous cells in the array.

\item Any one pattern may appear zero or more times in an alignment.

\item Where a pattern appears two or more times in an alignment, no symbol in one appearance of the pattern should ever be shown as matching the same symbol in another appearance of the pattern.

\item Any symbol in one pattern may be placed in the same column as any other symbol from the same pattern or another pattern, providing {\it order constraints} are not violated.
\end{itemize}

For any alignment, {\bf order constraints} are preserved if the following statement is always true:

\begin{quote}
For any two rows in the alignment, A and B, and any four symbols, A$_1$ and A$_2$ in A, and B$_1$ and B$_2$ in B, if A$_1$ is in the same column as B$_1$, and if A$_2$ is in the same column as B$_2$, and if A$_2$ in A follows A$_1$ in A, then B$_2$ in B must follow B$_1$ in B. 
\end{quote}

\noindent This condition holds when the two rows contain two different patterns and also when the two rows contain two appearances of one pattern.

\subsection{`Mismatch'}

A {\bf mismatch} in an alignment occurs when, between two columns in
the alignment containing hits, or between one column containing hits
and the beginning or end of the alignment, there are no other columns
containing hits and there are two more columns containing single
symbols from two or more different patterns in Old.

\section{Evaluation of an alignment in terms of
compression}\label{appendix_B}

Section \ref{MA_compress} described in outline how, in the ICMAUS
scheme, an alignment is evaluated in terms of compression. This section
provides more detail.

As explained earlier, an alignment and its unification is interpreted
as a means of encoding New or part of New in terms of patterns in Old.
If New (the sentence to be parsed) or part of New is matched by a
pattern in Old (the grammar) then code symbols from that pattern may be
used as an abbreviated description of that part of New.

By contrast with standard compression methods, each code serves a dual
role: to identify the corresponding pattern uniquely within the
grammar, and to mark the left and right ends of the pattern. For
present purposes, the second role is required to remove the ambiguity
which would otherwise exist about left-to-right sequencing of symbols
in alignments.

As noted in Section \ref{MA_compress}, the coding principle may be
applied through two or more `levels' so that the symbols which encode a
sequence of two or more patterns at one level may themselves be
recognised as an instance of a recurrent pattern which has its own code
at the next higher level. Examples will be seen below.

A key point in this connection is that a recurrent pattern may be
discontinuous in the sense that the symbols in the pattern are not
necessarily contiguous as they appear in any or all of its occurrences.
In other words, a recurrent pattern may appear as a subsequence within
larger patterns. Thus, for example, a sequence of symbols like `A B C D
E F' may be recognised as a recurrent pattern within a set of instances
which includes patterns like `P A B Q C R D E F S', `A L B C D M N E F
O P', `X A B C D Y E F Z' and so on.

In what follows (Appendices \ref{individual_symbols} to
\ref{sizes_of_gaps}), I shall first give an informal
explanation of the method of calculating the compression associated
with any alignment using the example shown in Figure \ref{basic1}. Then
the principles embodied in the method are discussed in Appendix
\ref{discussion} and a formal summary of the method is presented in
Appendix \ref{summary_of_method}.

\subsection{Encoding individual symbols}\label{individual_symbols}

The simplest way to encode individual symbols in the sentence and the
grammar is with a `block' code using a fixed number of bits for each
symbol. In the grammar in Figure \ref{grammar_2}, there are 24 symbol
types so the minimum number of bits required for each symbol is
$\lceil log_2 24\rceil = 5$ bits per symbol.

In fact, the SP61 model (described in Section \ref{the_SP61_model})
uses variable-length codes for symbols, assigned in accordance with the
Shannon-Fano-Elias (S-F-E) coding scheme (described by \cite{r11}) so
that the shortest codes represent the most frequent symbols and {\it vice
versa}.

Notice that the number of bits required for each symbol is entirely
independent of the number of characters in the name of the symbol as it
is shown in the examples. Names of symbols are chosen purely for their
mnemonic value and to aid comprehension.

There are many variations and refinements that may be made at this
level but, in general, the choice of coding system for individual
symbols is not critical for the principles to be described below where
the focus of interest is the exploitation of redundancy which may be
attributed to sequences of two or more symbols rather than any
redundancy attributed to unbalanced frequencies of individual symbols.

For reasons which are given in Section \ref{decompression_by_compression}
connected with the decoding of information, the code for each symbol
has two different sizes (in bits): a `minimum cost' which is the
theoretical minimum number of bits needed to represent that symbol
according to the S-F-E calculations, and an `actual cost' which is the
(larger) number of bits that are needed to allow robust decoding of
information as well as encoding.

In the following informal description of the encoding principles, the
distinction between the `minimum cost' and the `actual cost' of each
symbol is not important and will be ignored. For the sake of simplicity
in this presentation, it will be assumed that all symbols are encoded
with the same number of bits so that `one symbol' can be treated as the
minimum unit of information.

\subsection{Encoding words}

As explained in Section \ref{MA_compress}, a word like `t h i s' in the
grammar shown in Figure \ref{grammar_2} may be encoded as `D 0 \#D', In
a similar way, the word `l o v e s' may be encoded as `V 0 \#V' and
likewise for the other words. In all cases except `b o y', there is a
modest saving of one or two symbols for each word.

\subsection{Encoding phrases}\label{encoding_phrases}

Consider the phrase `t h i s b o y'. If this were encoded with a code
pattern for each word, the result would be `D 0 \#D N 1 \#N' which is
only one symbol smaller than the original. However, we can encode the
phrase with fewer symbols by taking advantage of the fact that the
sequence `D 0 \#D N 1 \#N' has a subsequence, `D \#D N \#N', which is a
substring within the pattern `NP D \#D N \#N \#NP' in the grammar. Notice
that the sequence `D \#D N \#N' is discontinuous within the sequence `D 0
\#D N 1 \#N' in the sense described earlier.

Since the `noun phrase' pattern `NP D \#D N \#N \#NP' is in the grammar,
we may replace the substring, `D \#D N \#N', by the `code' sequence `NP
\#NP'. But then, to encode the two words within the noun phrase (`t h i
s' and `b o y'), we must add the symbols, `0' and `1' from `D 0 \#D N 1
\#N' so that the final coded sequence is `NP 0 1 \#NP'.

Notice how the symbols `NP' and `\#NP' in the code pattern `NP 0 1 \#NP'
serve as a disambiguating context so that the symbol `0' identifies the
pattern `D 0 t h i s \#D' and the symbol `1' identifies the pattern `N 1
b o y \#N'.  The overall cost of the code pattern `NP 0 1 \#NP' is 4
symbols compared with the original 7 symbols in `t h i s b o y' - a
saving of 3 symbols. In a similar way, the phrase `t h a t g i r l' may
be encoded as `NP 1 0 \#NP' which is 4 symbols smaller than the
original.

\subsection{Encoding the sentence}\label{encoding_the_sentence}

Given the two noun phrases in their encoded forms (`NP 0 1 \#NP' for `t
h i s b o y' and `NP 1 0 \#NP' for `t h a t g i r l') and the encoding
of `l o v e s' as `V 0 \#V', the whole sentence may be encoded as `NP 0
1 \#NP V 0 \#V NP 1 0 \#NP'.

However, this sequence contains the subsequence `NP \#NP V \#V NP \#NP'
and this sequence is a substring within the `sentence' pattern `S NP
\#NP V \#V NP \#NP \#S' - and this pattern is in the grammar. So we may
replace the sequence `NP \#NP V \#V NP \#NP' by the `code' sequence `S
\#S'. To discriminate the words in this sentence we must add the symbols
`0 1 0 1 0' from the sequence `NP 0 1 \#NP V 0 \#V NP 1 0 \#NP'. The
overall result is an encoded representation of the sentence as:

\begin{center}
\begin{BVerbatim}
S 0 1 0 1 0 #S.
\end{BVerbatim}
\end{center}

The 7 symbols in this encoding of the sentence represents a substantial
compression compared with the 20 symbols in the unencoded sentence.

\subsection{Taking account of the sizes of gaps}\label{sizes_of_gaps}

The account of pattern matching and coding in Sections
\ref{encoding_phrases} and \ref{encoding_the_sentence} illustrates the
way in which `matching' in the proposed scheme embraces the matching of
subsequences (where the matched symbols need not be contiguous) as well
as the more traditional matching of coherent substrings (where the
matched symbols are always contiguous, one with the next).

In this connection, most people have a strong intuition that, where
there are gaps in matching, small gaps or no gaps are `better' than
large ones. It seems that our intuitions in this area can be justified
in terms of probability theory. A method, based on probability
principles, for making allowances for gaps has been developed and is
applied in the SP61 model. A brief outline of the method
and how it is applied is presented in Section \ref{summary_of_method},
below.

\subsection{Discussion}\label{discussion}

Each pattern expresses sequential redundancy in the data to be encoded
and this sequential redundancy can be exploited to reduce the number of
symbols which need to be written out explicitly. In the grammar shown
in Figure \ref{grammar_2}, each pattern for an individual word expresses the
sequential redundancy of the letters within that word; the pattern for
a noun phrase expresses the sequential redundancy of `determiner'
followed by `noun'; and the pattern for a sentence expresses the
sequential redundancy of the pattern: `noun phrase' followed by `verb'
followed by `noun phrase'.

Since this principle operates at all levels in the `hierarchy' of
patterns, many of the symbols at intermediate levels may be omitted
completely. A sentence may be specified with symbols marking the start
and end of the sentence pattern together with interpolated symbols
which discriminate amongst alternatives at lower levels.

Notice that these ideas are only applicable to alignments which can
`project' into a single sequence of symbols, as is the case with the
alignment shown in Figure \ref{basic1}. Any alignment like this:

\begin{center}
\begin{BVerbatim}
a x b             a b x
|   |   or this   | |
a y b             a b y
\end{BVerbatim}
\end{center}

\noindent where there is a `mismatch' of symbols, cannot be evaluated
in this way. For present purposes, any such alignment is excluded from
consideration. When the SP model is generalised to other areas such as
learning, it is intended that alignments like those just shown will be
evaluated alongside those which can project without mismatches.

The method that has been described illustrates the role of context in
the encoding of information. Any one symbol like `0' or `1' is
ambiguous in terms of the patterns in the grammar in Figure
\ref{grammar_2}. But in the context of the pattern `S 0 1 0 1 0 \#S'
and the same grammar, it is possible to assign each instance of `0' or
`1' unambiguously to one of the words in the grammar, giving the
sequence of words in the original sentence. It appears that ICMAUS
provides a mechanism for `decoding' the encoded form of the sentence,
as discussed in Section \ref{decoding_by_compression}.

\subsection{Summary of method for calculating the compression
associated with an alignment}\label{summary_of_method}

The proposed method of calculating the compression difference (CD)
associated with an alignment of patterns is summarised in more formal
terms here. This is the method embodied in the SP61 model (which is
described in Section \ref{the_SP61_model} and Appendix
\ref{appendix_C}).  The method is designed to calculate the compression
of New information or part of it (all or part of the sentence to be
parsed) which may be achieved by `encoding' New information in terms of
Old information (where Old information is the patterns of symbols
representing the grammar used in parsing). This CD is calculated as:

\[CD = B_N - B_E,\]

\noindent where $B_N$ is the number of bits required to represent the
hit symbols in New without any encoding (except S-F-E coding at the
level of single symbols), and $B_E$ is the number of bits required for
the encoding of those same symbols from New in terms of Old
information. How these values are calculated is described below.

\subsubsection{Information costs of
symbols}\label{information_costs_of_symbols}

If a simple block code is used for symbols, then the `minimum cost',
$M$, for each symbol is

\[M = \lceil log_2|S|\rceil \]

\noindent bits where $|S|$ is the number of symbol types in the
alphabet of symbol types ($S$) used throughout New and Old.

As previously noted, the value of $M$ for each symbol type (and thus
each individual symbol) is calculated in SP61 by the S-F-E method. For
any one symbol type, the input for this calculation is the frequency of
occurrence of the symbol type either measured directly or approximated
using this formula:

\[f_{st} = \sum_{i=1}^P (f_i \times o_i)\]

\noindent where $f_i$ is the (notional) frequency of the ith pattern in
the grammar (illustrated by the numbers on the right of Figure \ref{grammar_2}),
$o_i$ is the number of occurrences of the given symbol in the $i$th
pattern and $P$ is the number of patterns in the grammar.

Whichever way the value of $M$ is calculated, the `actual cost', $A$,
of each symbol is:

\[A = M \times c,\]

\noindent where $c$ is a factor whose size is not critical except that
$c > 1$.

\subsubsection{Calculation of $E$, the minimum number of bits required
for the encoding of a given pattern in Old}\label{bits_to_encode_pattern_old}

The calculation of $B_E$ for any alignment requires a value for the
`encoding cost', $E$, for each pattern from Old which appears in the
alignment.

Since there is a frequency of occurrence associated with each pattern
in any grammar, it is possible to calculate a theoretical minimum for
the value of $E$ for each pattern using the S-F-E method. However,
there is an alternative method of calculating $E$ which, for present
purposes, appears to be more useful and which has been adopted in the
SP61 model described in Section \ref{the_SP61_model}.

In summary, the alternative method is to calculate $E$ as

\[E = \sum_{i=1}^n D_i\]

\noindent where $D_i$ is the $M$ value for the $i$th symbol in a
subsequence of $n$ `discrimination' symbols within the given pattern
which identifies the pattern uniquely amongst the patterns in the
grammar without over-specifying the pattern.

Ideally, the discrimination symbols for a pattern would be whatever
subsequence of the pattern was most distinctive of the pattern,
regardless of the position of the symbols within the pattern. However,
in the SP61 model, two constraints have been imposed:

\begin{itemize}

\item The simplifying assumption has been made that the discrimination
symbols are the smallest substring of one or more symbols starting at
the beginning of the pattern which enables the pattern to be identified
uniquely within the grammar. For any pattern, it is easy to discover
what this substring is by a process of systematic comparison of
candidate substrings with corresponding symbols in other patterns in
the grammar.

Although a constrained subsequence of symbols is used in calculating
the value of $E$ for the pattern, this does not mean that a pattern can
only ever be recognised by those symbols and no others. In the SP61
model, a pattern can be fully or partially recognised by any
subsequence of its symbols.

\item Whenever a pattern ends in a `termination' symbol (a symbol
whose first character is the hash character (`\#')), this symbol is
added to the set of discrimination symbols for the pattern if it is not
otherwise there.

\end{itemize}

\subsubsection{Calculation of $B_N$ (the number of bits required to
represent hit symbols from New in `raw' form)}

For any one alignment, $B_N$ is calculated as:

\[B_N = \sum_{i=1}^h A_i\]

\noindent where $A_i$ is the `actual cost' of the symbol corresponding
to the $i$th hit in a sequence of hits, $H_1 ... H_h$, with an
adjustment to be described in the next paragraph. The hit sequence $H_1
... H_h$ comprises the hits between symbols in New and symbols in
patterns in Old. The symbols from New in this hit sequence are a
subsequence of the sequence $N_1 ... N_n$, which is the pattern in
New.

\subsubsection{Allowing for gaps}\label{allowing_for_gaps}

Before the formula, above, is applied, the value of each $A_i$ is adjusted
to take account of any `gap' which may exist between the given hit and
any previous hits in the sequence of hits between New and patterns in
Old. For this purpose, the alignment is treated as if it were two
sequences of symbols: the sequence of symbols which is New (the
sentence being parsed) and the sequence of symbols which is the
projection of the alignment into a single sequence.

As indicated above, there is insufficient space to present fully the
method of allowing for gaps. In outline, it is based on an analogy with
the rolling of two $A$-sided dice, where $A$ is the size of the alphabet
used in New and Old. The sequence of rolls of one die corresponds with
the sequence of symbols in New and the sequence of rolls of the other
die corresponds with the sequence of symbols in the projection. The
method is based closely on the method described in \cite{r20} for
calculating probabilities of various contingencies in problems of this
type.

For the symbol corresponding to the $i$th hit in the sequence $H_1 ...
H_h$, the adjusted value of $A_i$ is calculated as:

\[A_i = a_i \times F_s\]

\noindent where $a_i$ is the actual cost of the symbol corresponding to
the $i$th hit in $H_1 ... H_h$, and $F_s$ is the $s$th entry in a table
of `scaling factors' which is calculated at the outset of processing.
The value of $F_1$ is always 1.  For each hit in $H_1 ... H_h$ after
the first, the variable $s$ (which represents the `span' between the
current hit in $H_1 ... H_h$ and the preceding hit) is calculated as:

\[s = (P_i - P_{i-1}) \times (C_i - C_{i-1})\]

\noindent where $P_i$ is the position in $N_1 ... N_n$ of the symbol
corresponding to the $i$th hit in $H_1 ... H_h$, $P_{i-1}$ is the
position in $N_1 ... N_n$ of the symbol corresponding to the ($i-1$)th
hit in $H_1 ... H_h$. $C_i$ and $C_{i-1}$ are the analogous positions
in the projection of the alignment into a single sequence - which means
that $C_i$ and $C_{i-1}$ represent columns in the alignment itself.

\subsubsection{Calculation of $B_E$ (the number of bits required to
encode the hit symbols from New)}\label{bits_to_encode_new}

For each new alignment, the value of $B_E$ is:

\[B_E = \sum_{i=2}^r E_i - S\]

\noindent where $E_i$ is the `encoding cost' of the Old pattern
appearing on one of $r$ rows of the alignment other than the top line
(where New appears) and $S$ is the saving in encoding costs arising
from the fact that some patterns in the alignment convey information
about the sequential arrangement of other patterns in the alignment or
the selection of other patterns in the alignment where alternatives are
possible in a given context.

The `encoding cost' of any pattern is the value of $E$ for that
pattern, calculated as described in Appendix
\ref{bits_to_encode_pattern_old}. Notice that if any pattern appears two
or more times in the alignment, its encoding cost is added a
corresponding number of times to the sum of encoding costs.

The calculation of $B_E$ depends on three main ideas:

\begin{itemize}

\item As previously noted, a pattern may be fully or
partially recognised by any subsequence of the pattern. In other words,
it is not necessary to use the specific symbols which were used in
calculating the value of $E$ for that pattern. As a general rule when the
grammar is largely free of redundancy, if the $M$ values of the relevant
symbols (adjusted for gaps - see next) add up to the value of $E$ then
that subsequence of symbols will identify the pattern uniquely amongst
the other patterns in the grammar. Where there is redundancy in the
grammar, more bits may be needed to achieve unique identification of a
pattern.

\item If there are gaps in a sequence of hits, information values must
be reduced in accordance with the rules used in calculating the value
of $B_N$ (Appendix \ref{allowing_for_gaps}).

\item For present purposes, there is nothing to be gained by
over-specifying a pattern. If one pattern matches a second pattern by
the minimum number of symbols needed to achieve unique identification
of that second pattern, then the saving in encoding costs from this
source is maximal.  Any additional hits between the two patterns do not
give any additional saving in encoding costs.

\end{itemize}

$B_E$ is calculated in the following way:

\begin{enumerate}

\item For each row ($R$) in the alignment corresponding to a pattern from
Old, create a variable ($V$) containing the value of $E$ for the pattern in
that row.

\item Traverse the alignment from left to right examining the columns
containing two or more symbols (including symbols in New). Any such
column is designated a `hit' column ($C_H$).

\item For each $C_H$ which contains two or more symbols from patterns in
Old (which we may designate $C_{HO}$), examine each row which has a hit
symbol from Old in the column (designated $R_{HO}$). For this symbol,
calculate $M_A$, an `adjusted' value of $M$ for the symbol, taking account
of any gap which may exist between the given $C_{HO}$ and any previous $C_H$.
The method of making the adjustment is the same as is used for
calculating the value of $B_N$ (Section \ref{allowing_for_gaps}) except
that, for each $R_{HO}$, the gaps (or spans) are measured as if all the rows
in the alignment except the given $R_{HO}$ is treated as if it were a single
pattern to which the pattern in the given RHO is aligned. As in the
calculation of $B_N$, it is assumed that there is no gap associated with
the first $C_H$ for any given pattern.

\item For each $C_{HO}$, examine each $R_{HO}$ and, amongst these rows, identify
the `leading' row, $R_{HOL}$, whose pattern starts furthest to the left in
the alignment (if there is a tie, make an arbitrary choice amongst the
ties). For example, in Figure \ref{basic1}, for either of the two columns which
contains a hit between `D' in `D 0 t h i s \#D' and `D' in `NP D \#D N \#N
\#NP', the $R_{HOL}$ is the one containing `NP D \#D N \#N \#NP' (row 7 in the
first case and row 2 in the second case); for either of the two columns
containing a hit between `NP' in `NP D \#D N \#N \#NP' and `NP' in `S NP
\#NP V \#V NP \#NP \#S', the $R_{HOL}$ is the row containing `S NP \#NP V \#V NP
\#NP \#S' (row 5 in both cases).

\item For each $C_{HO}$, consider, in turn, each $R_{HO}$, excluding
the $R_{HOL}$.  For each row considered, subtract the value of $M_A$ from
the value of $V$ for that row. If the new value of $V$ is less than $0$, $V$ is
set to $0$ and no further subtraction from that instance of $V$ is
allowed.

\item When all relevant columns have been examined and the values of
the V variables have been reduced, calculate

\[B_E = \sum_{i=2}^r V_i\]

where $r$ is the number of rows in the alignment and the summation
excludes the top line (which contains New).

\end{enumerate}

The rationale for this method of calculating $B_E$ is that it gives us
the sum of the $E$ values of the patterns from Old corresponding to
each row of the alignment after the first, with a reduction for hits
between those patterns (with an adjustment for gaps as outlined
above).

The reason for reducing the value of $B_E$ when there are hits between
patterns in Old is that any such hit reflects a degree of `coverage' of
one pattern from Old by another such pattern. To the extent that one
pattern provides information that also exists in another pattern there
is a reduced need for the second pattern to be identified in the
encoding. In the extreme case, where two patterns are identical, only
one of them need be identified in the encoding.  As indicated above,
any saving in encoding costs resulting from the coverage of one or more
patterns by another cannot exceed the E value for each pattern - any
additional hits are `wasted'. Hence, the $V$ value for any row cannot be
reduced below $0$.

In the method described above, the `leading' row for any one column
($R_{HOL}$) is regarded as the row with which the other symbols in the
column are unified. Hence, for the given column, this is the row where
the $V$ value is not reduced by the value of $M_A$. Intuitively, the
left-to-right bias in the definition of `leading row' is less
theoretically `clean' than if all concepts were entirely symmetrical
between left and right directions in the alignment. However, the
concepts as described are the best to have been found so far and seem
to work quite well.

\section{The organisation and operation of the SP61 model}\label{appendix_C}

Figure \ref{SP61_structure} presents a high level view of the
organisation of the SP61 model using pseudocode while Figures
\ref{compress_structure_1} and \ref{compress_structure_2} show, with
pseudocode, the first and second parts of the {\it compress()} function
within the model. The text below describes how the model works together
with details of its organisation that are not included in the
pseudocode.

\begin{figure}[b!hpt]
\centering
\begin{BVerbatim}
main()
{
     1 Read the rules of the grammar, each one with a frequency
          of occurrence in a notional sample of the language,
          and store the patterns with their frequencies in Old.
     2 Read the sentence to be parsed and store it in New.
     3 Derive a frequency for each symbol in the grammar
          (as described in Appendix B).
     4 Using the frequencies of the symbols with the method,
          assign to each symbol in New and Old a number
          of bits representing the `minimum' information `cost'
          of that symbol. Also, calculate an `actual'
          information cost for each symbol.
     5 For each pattern in the grammar, calculate E, the
          minimum number of bits needed to encode that pattern.
     6 Select the sentence to be parsed and add it as the first
          `driving pattern' to an otherwise  empty list of
         driving patterns.
     7 while (new alignments are being formed) 
          compress ()
     8 Out of all the new alignments which have been formed,
          print the ones with the best CDs.
}
\end{BVerbatim}
\caption{\small A high level view of the organisation of the SP61 model.}
\label{SP61_structure}
\normalsize
\end{figure}

\begin{figure}[b!hpt]
\centering
\begin{BVerbatim}
compress()
{
     1 Clear the `hit structure' (described in the text).
     2 while (there are driving patterns that have not
          yet been processed)
     {
          2.1 Select the first or next driving pattern
               in the set of driving patterns.
          2.2 while (there are more symbols in the
               current driving pattern)
          {
               2.2.1 Working left to right through the
                    current driving pattern, select the
                    first or next symbol in the pattern.
               2.2.2 `Broadcast' this symbol to make a
                    yes/no match with every symbol in the
                    `target patterns' in Old.
               2.2.3 Record each positive match (hit) in a
                    `hit structure' (as described in the
                    text). As more symbols are broadcast,
                    the hit structure builds up a record
                    of sequences of hits between the
                    driving pattern and the several target
                    patterns in Old. As each hit sequence
                    is extended, the compression score of
                    the corresponding alignment is
                    estimated using a `cheap to compute'
                    method of estimation.
               2.2.4 If the space allocated for the hit
                    structure is filled at any time, the
                    system `purges' the worst hit sequences
                    from the hit structure to release more
                    space. The selection uses the estimates
                    of compression scores assigned to each
                    hit sequence in Step 2.2.3.
          }
     }
\end{BVerbatim}
\caption{\small First part of the {\it compress()} function of the SP61 model.}
\label{compress_structure_1}
\normalsize
\end{figure}

\begin{figure}[b!hpt]
\centering
\begin{BVerbatim}
     3 For each hit sequences which has an estimated
          compression score above some threshold value
          and which will `project' into a single
          sequence (as described in the text), convert
          the hit sequence into the corresponding
          alignment. Discard this alignment if it is
          identical with any alignment already in Old.
          Otherwise, compute the compression score using
          the method described in Appendix C, print
          the new alignment and add it to Old. If
          no new alignments are formed, quit the
          compress() function.
     4 Excluding the original patterns in Old, examine
          all the alignments that have been added to Old
          since the beginning of processing and choose
          a subset of these alignments using the method
          described in the text. Remove from Old all the
          alignments which have not been selected. The
          original patterns are never removed from Old.
     5 Clear the list of driving patterns and then, using
          the same method as is used in 4 but (usually)
          with a more restrictive parameter, select a
          subset of the alignments remaining in Old and
          add references to those alignments to the list
          of driving patterns (these patterns are not
          removed from Old and may therefore also be
          target patterns on the next cycle).
}
\end{BVerbatim}
\caption{\small Second part of the {\it compress()} function
of the SP61 model.}
\label{compress_structure_2}
\normalsize
\end{figure}

\subsection{Preliminary processing}

\subsubsection{Calculation of the information cost of each symbol}

As was described in Section \ref{representing_grammar}, each rule in
the grammar has an associated frequency of occurrence in (a `good'
parsing of) some notional sample of the language. In Step 3 of {\it
main()} in Figure \ref{SP61_structure}, the model derives the frequency
of occurrence of each symbol type as described in Appendix
\ref{information_costs_of_symbols}.

These frequencies are then used (in Step 4 of {\it main()}) to calculate the
minimum number of bits needed to represent each symbol type using the
S-F-E coding scheme (see \cite{r11}), as described in Appendix
\ref{information_costs_of_symbols}. The resulting sizes for each
symbol type are then assigned as `minimum cost' sizes to corresponding
symbols in New and Old. Each symbol in New and Old is also given an
`actual cost' which is the minimum cost increased by an arbitrary
factor, rounded up to ensure that the actual cost is at least one bit
larger than the minimum cost (see Section
\ref{information_costs_of_symbols}).

\subsubsection{Establishing the encoding cost of each pattern in Old}

In Step 5 of {\it main()} in Figure \ref{SP61_structure}, each
pattern in the grammar is assigned a minimum number of bits required to
discriminate the pattern from other patterns in the grammar using
frequencies of the patterns with the S-F-E method, as was outlined in
Section \ref{bits_to_encode_pattern_old}.

\subsection{Building the `hit structure' (step 2 of the {\it
compress()} function in Figure \ref{compress_structure_1})}

The {\it compress()} function shown in Figures
\ref{compress_structure_1} and \ref{compress_structure_2} is the heart
of the SP61 model.  This subsection and the ones that follow supplement
the description in the figure.

As can be seen from the figure and inferred from the outline
description in Section \ref{the_SP61_model}, the {\it compress()}
function is applied iteratively. On the first cycle, the `driving'
pattern is simply the sentence to be parsed. On subsequent cycles, the
list of driving patterns is a subset of the alignments formed in
preceding cycles. Iteration stops when no new alignments can be found
which satisfy conditions described below.

\subsubsection{Fuzzy matching of one pattern with another}

Step 2 of {\raggedright the {\it compress()} function is based on the central
process in SP21 \cite{r39}, a process which is related to dynamic
programming (DP, \cite{r31}) and is designed to find `fuzzy' matches
which are `good' between one `driving' pattern and one or more `target'
patterns.  In this context, a `fuzzy' match is one where only a
subsequence of the symbols in one pattern need match the symbols in the
other pattern and {\it vice versa}.

}

The technique is to `broadcast' each symbol in the driving pattern to
make a yes/no match with each symbol in the set of target patterns and
to record sequences of hits in a `hits structure'.
Each sequence of hits (termed a hit sequence) represents an alignment
between the driving pattern and one of the target patterns.

As is described in \cite{r39}, the hit structure has the form of a
list-processing tree with each node representing a hit and each path
from the root to a leaf node representing a sequence of hits.

\subsubsection{No one instance of a symbol should ever be matched with
itself}\label{no_self_matching}

Since driving patterns can also be target patterns, any one pattern may
be aligned with itself. That being so, a check is made to ensure that
no instance of a symbol is ever matched against itself (see Section
\ref{varieties_of_MA}). Obviously, any such match would be meaningless
in terms of the identification of redundancy.

Since any symbol in the driving pattern and any symbol in the target
pattern may have been derived by the unification of two or more other
symbols, a check is also made to exclude all hits where the set of
symbols from which one of the hit symbols was derived has one or more
symbols in common with the set of symbols from which the other hit
symbol was derived. In short, while any given pattern from the grammar
may appear two or more times in one alignment, no symbol in any of the
original patterns in Old ever appears in the same column as itself in
any alignment.

\subsubsection{The order of symbols in New must be
preserved}\label{preserve_symbol_order_in_new}

As the matching process has been described so far, it would be entirely
possible for the system to align a pattern like `NP D 1 t h a t \#D N 1
g i r l \#N \#NP' in the example considered earlier with the first `NP
\#NP' in a pattern like `NP \#NP V 0 l o v e s \#V NP \#NP \#S' from the
same example and to align `NP D 0 t h i s \#D N 0 b o y \#D \#NP' with the
second `NP \#NP'. To avoid the formation of alignments like this which
violate the order of the symbols in New, the system makes checks to
ensure, at all stages, that the order of the symbols in New is
honoured.

\subsubsection{Estimation of compression scores}

While the hit structure is being built, the compression score for the
alignment corresponding to each hit sequence may be calculated at every
stage but only at the cost of a lot of processing which would slow the
model down. Consequently, a simple method of estimating the compression
score is used in Step 2.2.3 of Figure \ref{compress_structure_1} which is
computationally `cheap'. Although it gives results which do not
correspond exactly with the values calculated using the formulae
presented in Appendix \ref{appendix_B}, the differences
appear not to be critical for the purposes of purging the hit structure
(Step 2.2.4 in Figure \ref{compress_structure_1}, Appendix
\ref{purging_of_hit_structure}) or determining the threshold for
converting hit sequences into alignments (Step 3 in
Figure \ref{SP61_structure}, Appendix \ref{creating_alignmments}).

\subsubsection{Purging the hit
structure}\label{purging_of_hit_structure}

If the space allocated to the hit structure is exhausted at any time,
the hit structure is `purged' or, more literally, `pruned' to remove
branches corresponding to the worst 50\% of the hit sequences (where the
meaning of `worst' is determined using the estimates of compression
scores calculated in Step 2.2.3 of the {\it compress()} function). In this
way, space is released in which new sequences of hits can be stored.

\subsubsection{Distinctive features of the technique}

The technique of recording hits in a tree using list processing,
coupled with the mechanism for purging the hit structure whenever the
available space is filled, is probably the most important difference
between the SP21 technique for finding partial matches and the more
traditional kinds of DP. In the SP21/SP61 technique:

\begin{itemize}

\item Both strings being compared can be arbitrarily long.

\item The `depth' of searching can be controlled by varying the space
available for the hit structure: larger spaces give better results than
smaller ones.

\item Unlike standard DP algorithms, the system delivers a set of
alternative alignments between two sequences rather than a single
`best' alignment.

\end{itemize}

\subsection{Building, scoring and selection of
alignments}\label{creating_alignmments}

\subsubsection{Building alignments and scoring them (step 3 of the
{\it compress()} function)}

When the hit structure for a set of driving patterns has been built,
the best hit sequences are converted into the corresponding alignments,
excluding all alignments which will not `project' on to a single
sequence (as described in Section \ref{parsing_as_alignment} and
Appendix \ref{discussion}) and excluding alignments as described in
\ref{later_selection_of_alignments}.

The process of converting a hit sequence into an alignment achieves two
things: it creates a one-dimensional sequence of symbols which is a
unification of the driving pattern or patterns with the target pattern
and it creates a two-dimensional array representing the alignment
itself. For each alignment, the array occupies a portion of memory of
exactly the right size, allocated dynamically at the time the alignment
is formed.

The one-dimensional sequence may enter into matching and unification in
later iterations of the {\it compress()} function, while the two-dimensional
array allows the full structure of the alignment to be seen and can be
used in later checks to ensure that no instance of a symbol is ever
matched with itself (Section \ref{no_self_matching}) and to ensure that
the order of symbols in New is not violated (Section
\ref{preserve_symbol_order_in_new}).

From time to time, identical alignments are formed via different
routes. The program checks each of the newly-formed alignments against
alignments already formed. Any alignment which duplicates one already
formed is discarded. The process of comparing alignments is indifferent
to the order (from top to bottom) in which patterns appear in the
alignment (cf. Section \ref{parsing_as_alignment}, above).

Every new alignment which survives the several hurdles is added to Old
and its CD is computed using the method and formulae described in
Appendix \ref{appendix_B}.

\subsubsection{Selection of alignments: a quota for each hit symbol in
New}\label{later_selection_of_alignments}

Apart from purging the hit structure when space is exhausted, the
main way in which the SP61 model narrows its search space is a
two-fold selection of alignments at the end of every cycle of the
{\it compress()} function:

\begin{itemize}

\item Excluding all the original patterns in Old, the program examines
the alignments which have been added to Old since the start of
processing and selects a subset by a method to be described. All the
other alignments are removed from Old and discarded.

\item Using the same method, the program selects a subset of the
alignments which remain in Old to be used as driving patterns on the
next cycle. These alignments are not removed from Old so they may also
function as target patterns.

\end{itemize}

At first sight it seems natural to select alignments purely on the
basis of their compression scores. However, it can easily happen that,
at intermediate stages in processing, the best alignments are trivial
variations of each other and involve the same subset of the symbols
from New. If selection is made by choosing alignments with a CD above a
certain threshold, the alignments which are chosen may all involve the
same subset of the symbols in New, while other alignments, containing
symbols from other parts of New, may be lost. If this happens, the
model cannot ever build an alignment which contains all or most the
symbols in New and may thus never find the `correct' answer.

A solution to this problem which seems to work well is to make
selections in relation to the symbols in New which appear in the
alignments. Each symbol in New is assigned a `quota' (the same for all
symbols) and, for each symbol, the best alignments up to the quota are
identified. Any alignment which appears in one or more of the quotas is
preserved. All other alignments are purged.  The merit of this
technique is that it can `protect' any alignment which is the best
alignment for a given subsequence of the symbols in New (or is second
or third best etc) but which may, nevertheless, have a relatively low
CD compared with other alignments in Old.

\subsection{Processing New in Stages}\label{windows}

A feature of the SP61 model that, to avoid clutter, has been omitted
from Figure \ref{SP61_structure} is that New may be divided into
`windows' of any fixed size (determined by the user) and the model can
be set to process New in stages, one window at a time, from left to
right. This feature of the model was introduced for two reasons:

\begin{itemize}

\item It seems to bring the model closer to the way people seem to operate,
processing sentences stage by stage as they are heard or read, not waiting
until the whole of a sentence has been seen before attempting to analyse
it.

\item Since it is possible to discard all but the best intermediate results
at the end of each window, this mode of processing has the advantage of
reducing peak demands for storage of information and it also has the effect
of reducing the size of the search space.

\end{itemize}


\begin{thebibliography}{}
%
\bibitem[Abney 97]{r1} Abney, S. P.: ``Stochastic attribute-value
grammars''; {\it Computational Linguistics}, 23, 4 (1997) 597-618.

\bibitem[Allison and Wallace 94]{r2} Allison, L. and Wallace, C. S.:
``The posterior probability distribution of alignments and its
application to parameter estimation of evolutionary trees and to
optimization of multiple alignments''; {\it Journal of Molecular
Evolution}, 39 (1994) 418-430.

\bibitem[Allison {\it et al.} 92]{r3} Allison, L, Wallace, C. S. and
Yee, C. N.: ``Finite-state models in the alignment of macromolecules'';
{\it Journal of Molecular Evolution}, 35 (1992) 77-89.

\bibitem[Barton 90]{r5} Barton G. J.: ``Protein Multiple Sequence
Alignment and Flexible Pattern Matching''; {\it Methods in Enzymology},
183 (1990) 403-428.

\bibitem[Belloti and Gammerman 96]{r6} Belloti, T. and Gammerman, A.;
``Experiments in solving analogy problems using Minimal Length
Encoding''; Presented at Applied Decision Technologies '95, Brunel
University, April 1995. Proceedings of Stream 1, Computational Learning
and Probabilistic Reasoning  (1996) 209-220).

\bibitem[Berger {\it et al.} 96]{r7} Berger, A. L., Della Pietra, S.
A.  and Della Pietra, V. J.: ``A maximum entropy approach to natural
language processing''; {\it Computational Linguistics}, 22, 1 (1996)
39-71.

\bibitem[Black {\it et al.} 93]{r8} Black, E., Garside, R. and Leech,
G. (Eds.):  {\it Statistically-driven computer grammars of English: the
IBM/Lancaster approach};  Rodopi, Amsterdam (1993).

\bibitem[Chan {\it et al.} 92]{r9} Chan, S. C., Wong, A. K. C. and
Chiu, D. K. Y. A.: ``Survey of Multiple Sequence Comparison Methods'';
{\it Bulletin of Mathematical Biology}, 54, 4 (1992) 563-598.

\bibitem[Cheeseman 90]{r10} Cheeseman, P.: ``On finding the most
probable model''. In J. Strager and P. Langley (Eds.) {\it
Computational models of scientific discovery and theory formation},
Chapter 3, Morgan Kaufmann, San Mateo, California, (1990) 73-95.

\bibitem[Chomsky 57]{r10a} Chomsky, N.: {\it Syntactic Structures};
Mouton, The Hague (1957).

\bibitem[Cover and Thomas 91]{r11} Cover, T. M. and Thomas, J. A.: {\it
Elements of Information Theory}; John Wiley, New York (1991).

\bibitem[Day and McMorris 92]{r12} Day, W. H. E. and McMorris, F. R.:
``Critical Comparison of Consensus Methods for Molecular Sequences'';
{\it Nucleic Acids Research}, 20, 5 (1992) 1093-1099.

\bibitem[Dreuth and Ruber 97]{r13} Dreuth, E. W. and Ruber B.:
``Context-dependent probability adaptation in speech understanding'';
{\it Computer Speech and Language}, 11 (1997) 225-252.

\bibitem[Felsenstein 81]{r14} Felsenstein, J.: ``Evolutionary trees
from DNA sequences: a maximum likelihood approach''; {\it Journal of
Molecular Evolution}, 17 (1981) 368-376.

\bibitem[Gammerman 91]{r15} Gammerman, A. J.: ``The representation and
manipulation of the algorithmic probability measure for problem
solving''; {\it Annals of Mathematics and Artificial Intelligence}, 4
(1991) 281-300.

\bibitem[Garside {\it et al.} 87]{r16} Garside, R., Leech, G. and
Sampson, G. (Eds.): {\it The Computational Analysis of English: A
Corpus-Based Approach}; Longman, London (1987).

\bibitem[Gazdar 89]{r17} Gazdar, G. and Mellish, C.: {\it Natural
Language Processing in Prolog}. Addison-Wesley, Wokingham (1989).

\bibitem[Hu {\it et al.} 97]{r18} Hu, J., Turin, W. and Brown, M. K.:
``Language modelling using stochastic automata with variable length
contexts''; {\it Computer Speech and Language}, 11 (1997) 1-6.

\bibitem[Li and Vitanyi 93]{r19} Li, M. and Vitanyi, P.: {\it An
Introduction to Kolmogorov Complexity and Its Applications}.
Springer-Verlag, New York (1993).

\bibitem[Lowry 89]{r20} Lowry, R.: {\it The Architecture of Chance};
Oxford University Press, Oxford (1989).

\bibitem[Lucke 95]{r21} Lucke, H: ``Bayesian belief networks as a tool
for stochastic parsing''; {\it Speech Communication}, 16 (1995)
89-118.

\bibitem[Pednault 91]{r22} Pednault, E. P. D.:
``Minimal-length encoding and inductive inference''. In G.
Piatetsky-Shapiro and W. J. Frawley (eds.), {\it Knowledge Discovery in
Databases}; MIT Press, Cambridge Mass (1991).

\bibitem[Pereira and Warren 80]{r23} Pereira, F. C. N. and Warren, D.
H. D.:  ``Definite Clause Grammars for language analysis - a survey of
the formalism and a comparison with augmented transition networks'';
{\it Artificial Intelligence}, 13 (1980) 231-278.

\bibitem[Reichert {\it et al.} 73]{r24} Reichert, T. A., Cohen, D. N.
and Wong, A. K. C.: ``An application of information theory to genetic
mutations and the matching of polypeptide sequences''; {\it Journal of
Theoretical Biology}, 42 (1973) 245-261.

\bibitem[Rissanen 78]{r25} Rissanen, J.: ``Modelling by the shortest
data description''; {\it Automatica-J., IFAC} 14 (1978) 465-471.

\bibitem[Solomonoff 64]{r26} Solomonoff, R. J.: ``A formal theory of
inductive inference, parts I and II''; {\it Information and Control}, 7
(1964) 1-22 and 224-254.

\bibitem[Storer 88]{r28} Storer, J. A.: {\it Data Compression: Methods
and Theory}; Computer Science Press, Rockville, Maryland (1988).

\bibitem[Takahashi and Sagayama 97]{r29} Takahashi, J. and Sagayama,
S.: ``Vector-field smoothed Bayesian learning for fast and incremental
speaker/telephone-channel adaptation''; {\it Computer Speech and
Language}, 11 (1997) 127-146.

\bibitem[Taylor 88]{r30} Taylor, W. R.: ``Pattern matching methods in
protein sequence comparison and structure prediction''; {\it Protein
Engineering}, 2, 2 (1988) 77-86.

\bibitem[Wagner and Fischer 74]{r31} Wagner, R. A. and Fischer, M. J.:
``The string-to-string correction problem''; {\it Journal of the ACM},
21, 1 (1974) 168-173.

\bibitem[Wallace and Boulton 68]{r32} Wallace, C. S. and Boulton, D.
M.: ``An information measure for classification''; {\it Computer
Journal}, 11, 2 (1968) 185-195.

{\raggedright

\bibitem[Wolff 00]{r32a} Wolff, J. G.:  
``Mathematics and logic as information compression by
multiple alignment, unification and search''; School of Informatics
Report, March 2000. A copy may be obtained from 
http://www.sees.bangor.ac.uk/\~gerry/sp\_summary.html\#maths\_logic.

\bibitem[Wolff 99a]{r33} Wolff, J. G.: ```Computing' as information
compression by multiple alignment, unification and search''; {\it
Journal of Universal Computer Science}, 5, 11 (1999a) 777-815. A copy
may be obtained from
http://www.jucs.org/jucs\_5\_11/computing\_as\_information\_compression.

\bibitem[Wolff 99b]{r34} Wolff, J. G.: ``Probabilistic
reasoning as information compression by multiple alignment, unification
and search:  an introduction and overview''; {\it Journal of Universal
Computer Science}, 5, 7 (1999b) 417-472. A copy may be obtained from:
http://www.jucs.org/jucs\_5\_7/probabilistic\_reasoning\_as\_information.
The three articles on which this
article is based may be obtained from http://www.iicm.edu/wolff/1998a,
b, c.

}

\bibitem[Wolff 98a]{r34a} Wolff, J. G.: ``Parsing as information compression
by multiple alignment, unification and search: SP52''; SEECS Report,
February 1998. A copy may be obtained from: http://www.iicm.edu/wolff/1998e.

\bibitem[Wolff 98b]{r34b} Wolff, J. G.: ``Parsing as information compression
by multiple alignment, unification and search: examples''; SEECS Report,
February 1998. A copy may be obtained from: http://www.iicm.edu/wolff/1998f.

{\raggedright

\bibitem[Wolff 98c]{r34c} Wolff, J. G.: ``Probabilistic
reasoning as information compression by multiple alignment,
unification and search''; SEECS Report, December 1998. A
copy may be obtained from:
http://www.sees.bangor.ac.uk/\~gerry/sp\_summary.html\#PrbRs.

}

\bibitem[Wolff 97]{r35} Wolff, J. G.: ``Causality, statistical learning
and multiple alignment''; Paper presented at the UNICOM Seminar and
Tutorial on Causal Models and Statistical Learning, London, March
1997.

\bibitem[Wolff 96]{r36} Wolff, J. G.: ``Learning and reasoning as
information compression by multiple alignment, unification and
search''; In: A. Gammerman (ed.), {\it Computational Learning and
Probabilistic Reasoning}, Wiley, Chichester (1996) 67-83. An earlier
version was presented at Applied Decision Technologies '95, Brunel
University, April 1995 (Proceedings of Stream 1, {\it Computational
Learning and Probabilistic Reasoning} 223-236).

\bibitem[Wolff 95a]{r37} Wolff, J. G.: ``Computing as compression: an
overview of the SP theory and system''; {\it New Generation Computing},
13 (1995) 187-214.

\bibitem[Wolff 95b]{r38} Wolff, J. G.: ``Computing as compression:
SP20''; {\it New Generation Computing} 13 (1995) 215-241.

\bibitem[Wolff 94a]{r39} Wolff, J. G.: ``A scaleable technique for
best-match retrieval of sequential information using metrics-guided
search''; {\it Journal of Information Science}, 20, 1 (1994a) 16-28.

\bibitem[Wolff 94b]{r40} Wolff, J. G.: ``Towards a new concept of
software''; {\it Software Engineering Journal}, 9, 1 (1994b) 27-38.

\bibitem[Wolff 94c]{r41} Wolff, J. G.: ``Computing and information
compression: a reply''; {\it AI Communications}, 7, 3/4 (1994c)
203-219.

\bibitem[Wolff 93]{r42} Wolff, J. G.: ``Computing, cognition and
information compression''; {\it AI Communications}, 6, 2 (1993)
107-127.

\bibitem[Wolff 91]{r43} Wolff, J. G.: {\it Towards a Theory of
Cognition and Computing}; Ellis Horwood, Chichester (1991).

\bibitem[Wolff 90]{r44} Wolff, J. G.: ``Simplicity and power: some
unifying ideas in computing''; {\it Computer Journal}, 33, 6 (1990)
518-534.

\bibitem[Wolff 88]{r45} Wolff, J. G.: ``Learning syntax and meanings
through optimization and distributional analysis''; In Y. Levy, I. M.
Schlesinger and M. D. S. Braine (Eds.), {\it Categories and Processes
in Language Acquisition}; Lawrence Erlbaum, Hillsdale, NJ (1988).
Reprinted in Chapter 2 of \cite{r43}.

\bibitem[Wolff 87]{r45a} Wolff, J. G.: ``Cognitive development as
optimisation''; In L. Bolc (Ed.), {\it Computational Models of
Learning}, Springer-Verlag, Heidelberg (1987) pp. 161-205.

\bibitem[Wolff 82]{r46} Wolff, J. G.: ``Language acquisition, data
compression and generalization''; {\it Language and Communication}, 2
(1982) 57-89. Reprinted in Chapter 3 of \cite{r43}.

\bibitem[Wu 97]{r47} Wu, D.: ``Stochastic inversion transduction
grammars and bilingual parsing of parallel corpora''; {\it
Computational Linguistics}, 23, 3 (1997) 377-403.

\end{thebibliography}
\end{document}